\documentclass{article}

\usepackage{microtype}
\usepackage{graphicx}
\usepackage{subcaption}
\usepackage{booktabs}
\usepackage{hyperref}
\usepackage{url}

\usepackage[preprint,nohyperref]{icml2026}

\usepackage{amsmath}
\usepackage{amssymb}
\usepackage{mathtools}
\usepackage{amsthm}
\usepackage{multirow}

\theoremstyle{plain}

\theoremstyle{definition}

\theoremstyle{remark}

\icmltitlerunning{ReflexGrad: Within-Episode Failure Recovery in LLM Agents}

\begin{document}

\twocolumn[
\icmltitle{ReflexGrad: Within-Episode Failure Recovery in LLM Agents via Progress-Gated Dual-Process Routing}
\pdfstringdefDisableCommands{\def\\{ }}

\icmlsetsymbol{equal}{*}

\begin{icmlauthorlist}
\icmlauthor{Ankush Kadu}{qpi}
\icmlauthor{Aswanth Krishnan}{qpi}
\end{icmlauthorlist}

\icmlaffiliation{qpi}{QpiAI}

\icmlcorrespondingauthor{Ankush Kadu}{ankush.k@qpiai.tech}

\icmlkeywords{LLM agents, inference-time learning, failure recovery, ALFWorld, dual-process architecture}

\vskip 0.05in
\centerline{\small \texttt{\{ankush.k,\, ashwanth.krishnan\}@qpiai.tech}}
\vskip 0.15in
]

\printAffiliationsAndNotice{}

\begin{abstract}
We present ReflexGrad, a dual-process architecture for within-episode failure recovery in LLM agents without demonstrations. When agents commit to a wrong approach early and exhaust the step budget, the post-failure trajectory contains the information to escape---but no published architecture acts on it within a single episode. ReflexGrad routes between a fast process (TextGrad-style continuous refinement every $k{=}3$ steps) and a slow process (Reflexion-style causal diagnosis when $m{=}5$ consecutive low-progress scores fire a routing gate). A deterministic priority merge keeps the natural-language policy coherent, and each slow activation emits three observable artifacts: a reproducible trigger, a causal diagnostic, and a verified fix. On ALFWorld 134 tasks, $n{=}10$ seeds, no demonstrations, ReflexGrad lifts Qwen-3-8B from $35.1\%$ to $75.4\%$ ($+40.3$pp), beating compute-matched 1-shot LATS by $+2.7$pp ($p{\approx}0.01$), ToT by $+5.7$pp ($p{<}10^{-4}$), and Self-Refine by $+6.7$pp ($p{<}10^{-5}$); on GPT-5 the lift is $46.3{\to}88.1\%$ ($+41.8$pp). The $1.5$pp cross-model difference is within seed noise ($p{\approx}0.13$), suggesting that the routing mechanism, rather than model scale, is the primary source of the gain. Code, prompts, per-seed logs, and sensitivity sweeps are released.\footnote{Accepted at the ICML 2026 Workshop on the Foundations of Deep Generative Models (FoGen). Code at \url{https://github.com/qpiai/reflexgrad}.}
\end{abstract}

\section{Introduction}

LLM agents fail on tasks they could solve. The agent commits to a wrong approach early, environment feedback is uninformative, minor variations repeat, and the step budget runs out~\citep{shinn2023reflexion, kim2025reflact}. Crucially, the information needed to escape the loop \emph{already exists} in the post-failure trajectory---but existing methods either delay recovery to the next trial or refine the wrong strategy locally.

The gap sits between two existing methods. Reflexion~\citep{shinn2023reflexion} performs strategic correction---verbal self-critique followed by a full retry---but delays recovery to the next trial and requires demonstrations to bootstrap. TextGrad~\citep{yuksekgonul2024textgrad} performs tactical correction---textual gradients that refine the policy within a single session---but the gradients are local and only sharpen a wrong strategy. Inference-time search methods (LATS~\citep{zhou2024lats}, ToT~\citep{yao2023tot}) widen the action space per step but never update the policy. The missing piece is a mechanism that starts with tactical refinement and \emph{escalates} to strategic correction when tactical fixes stop working---within a single episode.

ReflexGrad fills this gap. A \textbf{fast process} applies per-step textual refinement every $k{=}3$ steps, catching errors that a single gradient can fix. A \textbf{slow process} performs stall-triggered causal replanning when $m{=}5$ consecutive low-progress scores indicate the fast process has converged to a dead end. A deterministic priority merge ($\text{plan} \succ \text{gradient} \succ \text{base policy}$) keeps the natural-language policy coherent, and a cooldown phase protects plan execution from premature gradient interference. Each slow activation produces three observable artifacts: a reproducible trigger, a causal diagnostic, and a verified fix.

\paragraph{Contributions.}
\begin{itemize}\itemsep0pt
    \item \textbf{Architecture.} A progress-gated routing rule that combines per-step textual refinement with stall-triggered causal replanning, with a deterministic priority merge that keeps the natural-language policy coherent across overlapping updates.
    \item \textbf{Demo-free in-episode learning.} On ALFWorld 134 tasks ($n{=}10$ seeds, no demonstrations): $88.1\% \pm 2.0$ on GPT-5 and $75.4\% \pm 2.2$ on Qwen-3-8B (open-weight, 8B). Compute-matched on Qwen-3-8B, the demo-free architecture beats 1-shot LATS, ToT, and Self-Refine.
    \item \textbf{Robustness with falsifiable scope.} Sensitivity sweeps over three routing thresholds bound success within $84.3$--$88.1\%$. The $5.2$pp residual gap to 1-shot ReflAct~\citep{kim2025reflact} localizes to two categories (Heat, Examine) that require world knowledge a demonstration transfers directly; we do not claim parity.
\end{itemize}

We evaluate with cross-model ablation (GPT-5 and Qwen-3-8B~\citep{qwen2025}), compute-matched baselines, sensitivity analysis, step-budget scaling, and cross-domain probes on TextWorld~\citep{cote2018textworld} and OSWorld~\citep{xie2024osworld}. All compute-matched gaps are statistically separated at $p{<}0.05$ ($n{=}10$); the cross-model gain difference of $1.5$pp remains within seed noise ($p \approx 0.13$). Code, prompts, per-seed logs, and baseline implementations are released.

\begin{figure*}[!t]
\centering
\includegraphics[width=0.95\textwidth]{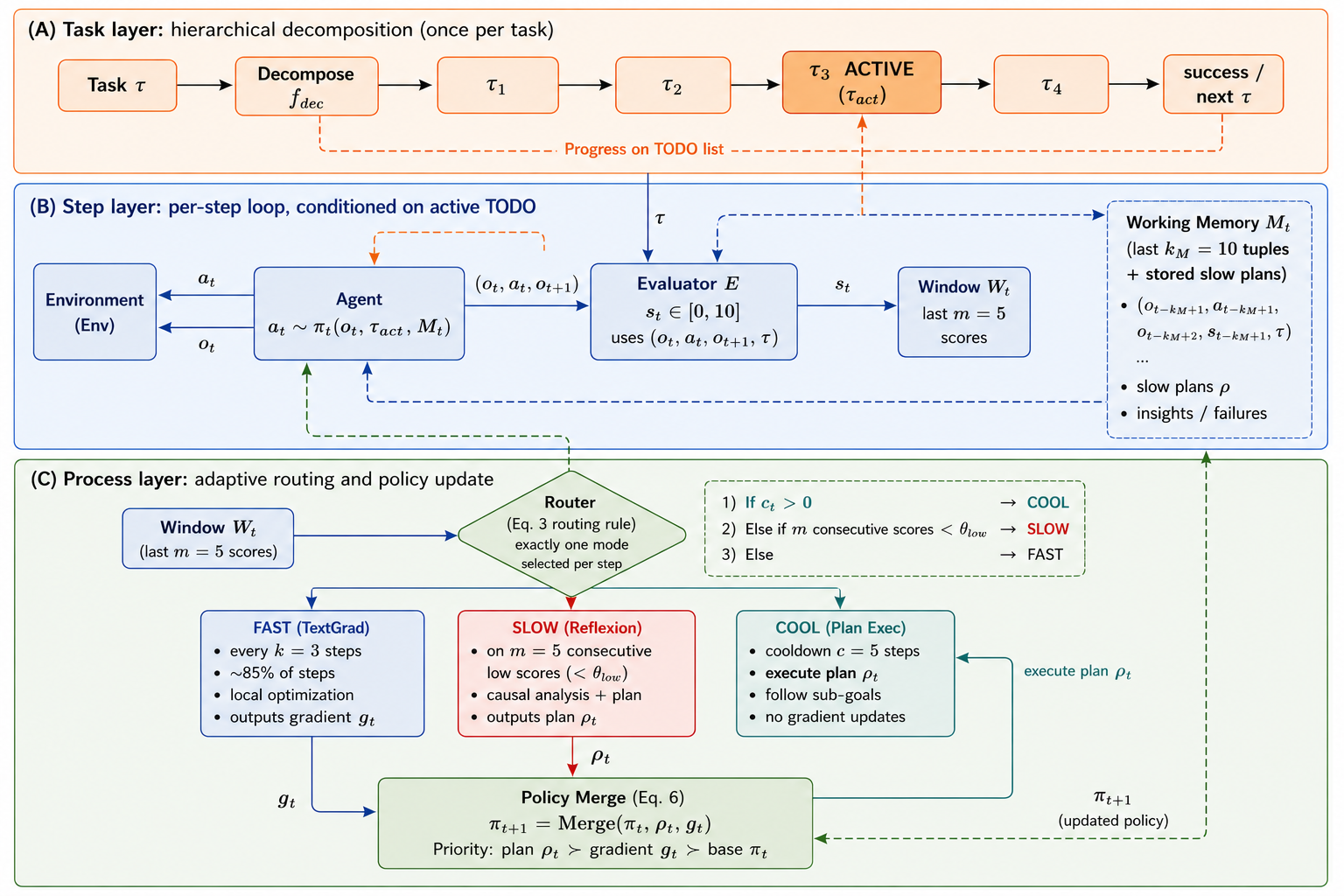}
\caption{\textbf{ReflexGrad bird's-eye architecture.} The agent acts on the environment; the evaluator $E$ scores each transition (Eq.~\ref{eq:score}) and scores accumulate in window $W_t$ (Eq.~\ref{eq:window}). The router (Eq.~\ref{eq:routing}) selects exactly one mode per step: FAST (TextGrad-style local refinement, every $k{=}3$ steps, $\sim 85\%$ of steps), SLOW (Reflexion-style causal reasoning on $m$ consecutive low scores), or COOL (plan execution under cooldown $c$, no gradient updates). The policy merge (Eq.~\ref{eq:merge}) combines updates under priority \emph{plan $\succ$ gradient $\succ$ base policy} and feeds $\pi_{t+1}$ back to the agent. \textbf{Failure-recovery artifacts} per slow activation: \textbf{(1)} the $m$-consecutive-low-score trigger; \textbf{(2)} the causal-trace diagnostic $d_t$; \textbf{(3)} the plan $\rho_t$ as the verified fix. Sub-stage internals: Fig.~\ref{fig:internals}; worked routing trace: Fig.~\ref{fig:trace}.}
\label{fig:architecture}
\end{figure*}

\begin{figure*}[!t]
\centering
\includegraphics[width=0.95\textwidth]{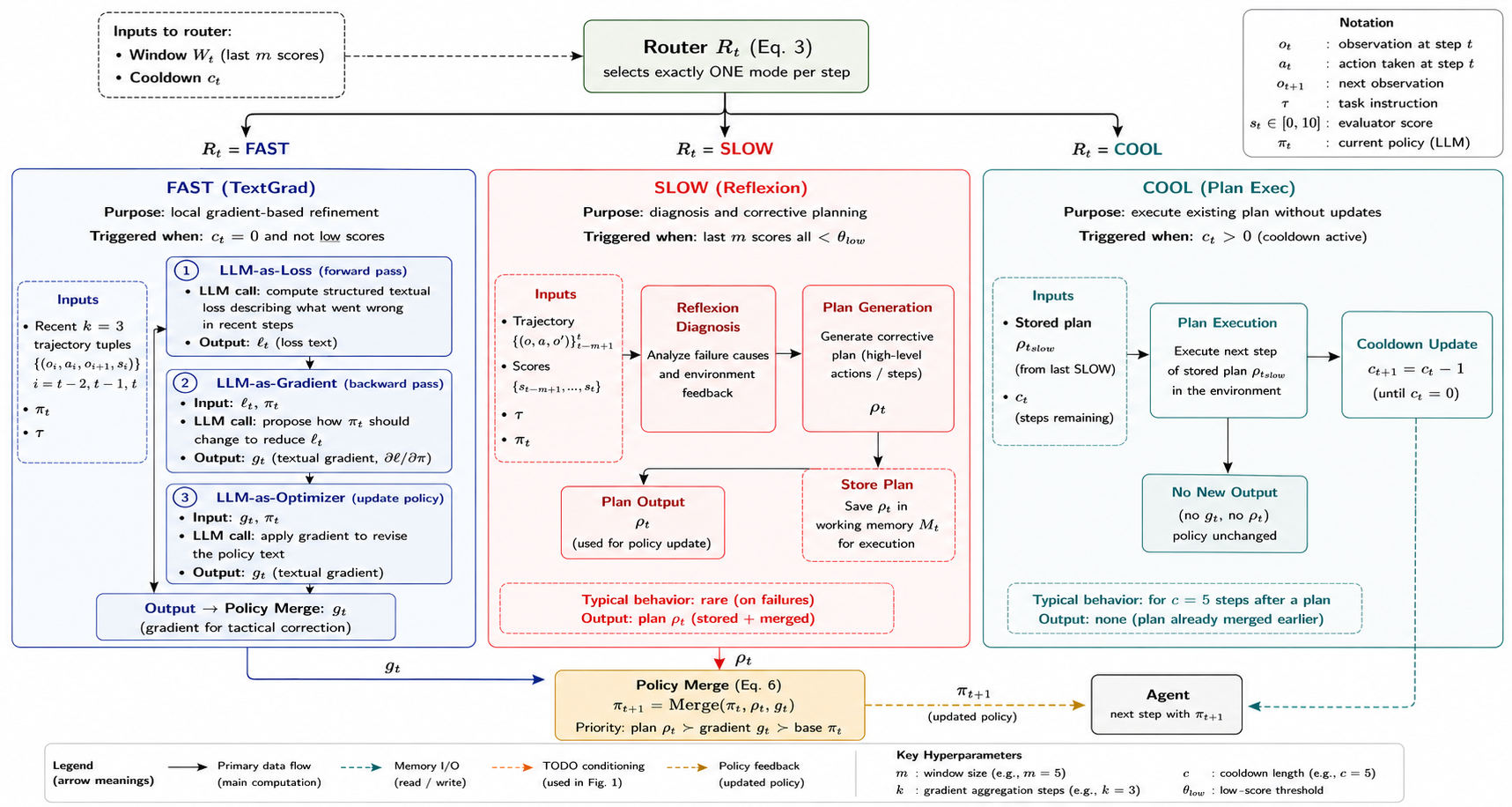}
\caption{\textbf{ReflexGrad sub-stage internals.} \textbf{Left (FAST process)}: TextGrad-style four-stage local optimization, executed every $k{=}3$ steps when the router selects FAST. Stages: \emph{recent context} $\to$ \emph{compute textual loss/gradient} $\ell_t, g_t$ $\to$ \emph{apply gradient} to policy $\to$ \emph{produce updated policy}. Quick gains, locally bounded, no causal reasoning. \textbf{Right (SLOW process)}: Reflexion-style four-stage causal reasoning, executed when consecutive low scores fire the routing gate. Stages: \emph{retrieve relevant memories} $\to$ \emph{reflect/diagnose} root cause $d_t$ $\to$ \emph{generate plan} $\rho_t$ as a sub-goal sequence $\to$ \emph{produce updated policy}. Slower, broader scope, simulates planning before acting. Both processes feed into the policy merge under priority $\rho_t \succ g_t \succ \pi_t$ (Eq.~\ref{eq:merge}). Exactly one process is selected per step by the router.}
\label{fig:internals}
\end{figure*}

\begin{figure*}[!t]
\centering
\includegraphics[width=0.95\textwidth]{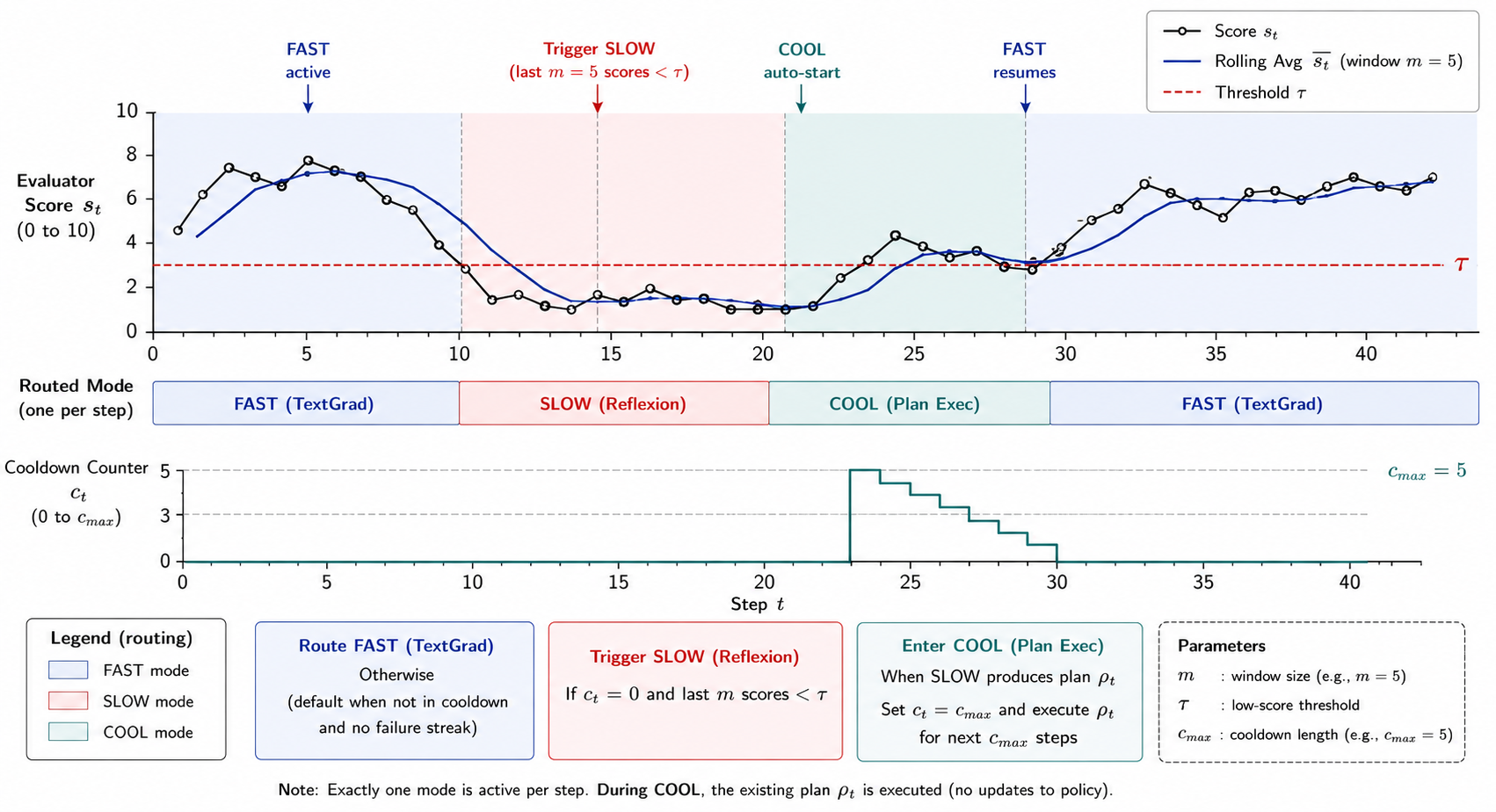}
\caption{\textbf{Routing over time.} Top: per-step performance signal $s_t$ from the evaluator $E$ (Eq.~\ref{eq:score}). Middle: router decision per step (FAST / SLOW / COOL) under the rule of Eq.~\ref{eq:routing}. Bottom: rolling window $W_t$ of last $m{=}5$ scores. \emph{High and stable scores} $\to$ FAST (gradient updates). \emph{Sustained low scores} (consecutive $\leq \theta_{\text{low}}$) $\to$ SLOW (reflection, plan generation). \emph{After reflection} $\to$ COOL for $c$ steps (execute plan without interference from gradients). After cooldown, FAST resumes. The router adapts continuously based on recent performance.}
\label{fig:trace}
\end{figure*}

\section{Background}

\paragraph{Reflexion.}
Reflexion learns across trials. The agent attempts a task, fails, writes a verbal self-reflection, then retries with the reflection in context. The published method uses 2 few-shot demonstrations and reaches $97\%$ on ALFWorld over 12 trials \citep{shinn2023reflexion}. Two properties block within-episode use: Reflexion requires demonstrations to bootstrap, and the reflection takes effect only on the next trial. Neither property fits within-episode recovery, where the step budget exhausts before the next trial begins.

\paragraph{TextGrad.}
TextGrad treats a natural-language string (the policy) as an optimizable parameter and computes textual gradients through LLM-based feedback \citep{yuksekgonul2024textgrad}. Each gradient step refines the policy from the immediate output, within a single session, with no demonstrations. The limitation is locality: a gradient adjusts the next action, not the strategy. When the strategy itself is wrong, the optimizer tunes the wrong policy more precisely.

\paragraph{The gap.}
Demo-free in-episode learning sits between these two methods. Reflexion's strategic correction needs the next trial; TextGrad's tactical correction stays inside the wrong strategy. No published method routes between both temporal scales within a single episode and merges their updates without conflict.

\section{Method}
\label{sec:method}

ReflexGrad has three components (Figure~\ref{fig:architecture}; Algorithm~\ref{alg:reflexgrad}): a fast process, a slow process, and a routing rule. We formalise the per-step setup, then describe each component.

\subsection{Setup and notation}

A task is a natural-language instruction $\tau$. Once per task, $f_{\text{dec}}(\tau,o_0)$ decomposes it into a TODO list $\mathcal{T}=\{\tau_1,\ldots,\tau_n\}$, $n \in [3,8]$, each with status $\sigma_i \in \{\text{pending}, \text{active}, \text{done}, \text{failed}\}$ and per-TODO state (attempts, last action, failure reasons) --- Fig.~\ref{fig:architecture}\,(A). The active TODO $\tau_{\text{act}}$ structures per-step decisions and acts as a checkpoint that prevents regression. At step $t$ the agent observes $o_t \in \mathcal{O}$, samples $a_t \sim \pi_t(o_t, \tau_{\text{act}}, M_t)$ where $\pi_t \in \Sigma^*$ is a natural-language policy and $M_t$ is a within-episode working memory (last $k_M{=}10$ tuples plus stored slow-process plans), and the environment returns $o_{t+1}$. Episodes terminate on success or step budget $T$. An LLM evaluator $E$ scores each transition:
\begin{equation}
s_t = E(o_t, a_t, o_{t+1}, \tau) \in [0, 10],
\label{eq:score}
\end{equation}
where higher values mean progress toward the goal. A rolling window holds the most recent $m$ scores:
\begin{equation}
W_t = (s_{t-m+1}, s_{t-m+2}, \ldots, s_t).
\label{eq:window}
\end{equation}
Reproducibility relies on the released $n{=}10$ seed list; per-call decoding uses each backend's default deterministic settings.

\subsection{Routing rule}

The router decides at each step whether to invoke the fast or slow process. The rule is deterministic in $W_t$ and a cooldown counter $c_t \geq 0$:
\begin{equation}
R_t =
\begin{cases}
\textsc{slow}, & \text{if } c_t = 0 \;\text{and}\; \forall i \in [t{-}m{+}1, t]:\; s_i < \theta_{\text{low}}, \\[2pt]
\textsc{cool}, & \text{if } c_t > 0, \\[2pt]
\textsc{fast}, & \text{otherwise.}
\end{cases}
\label{eq:routing}
\end{equation}
The slow case fires only when $m$ \emph{consecutive} scores fall below threshold $\theta_{\text{low}}$, not when their average is low. The conservative trigger suppresses spurious activation from a single noisy score. On activation, $c_t \leftarrow c{=}5$; the counter decrements each step, suppressing both processes during plan execution.

\paragraph{Why a progress-conditioned rule.} Fixed-cadence and random-gate routers ignore the structural signature the slow process targets: TextGrad's gradient is local and refines whatever policy it sees (including a wrong one), and the trace of a wrong strategy is consecutive low scores under continued local refinement --- exactly Equation~\ref{eq:routing}. The sensitivity sweep (Tab.~\ref{tab:sensitivity}) shows that the rule is robust across a range of threshold values, with the worst setting still producing $84.3\%$.

\subsection{Fast process: per-step textual refinement}

When $R_t = \textsc{fast}$ and $t \bmod k = 0$ ($k=3$), the system computes a textual gradient over the most recent $k$ tuples:
\begin{equation}
g_t = \mathrm{LLM}_{\text{grad}}\bigl(\pi_t,\; W_t[-k{:}],\; \{(o_i, a_i, o_{i+1})\}_{i=t-k+1}^{t}\bigr).
\label{eq:gradient}
\end{equation}
The gradient $g_t$ is a natural-language string proposing refinements to $\pi_t$, derived from the trajectory by the same loss--gradient--update template TextGrad uses for prompt optimization~\citep{yuksekgonul2024textgrad}. The fast process is local: each refinement reacts to the most recent $k$ steps, not the full episode. This delivers tactical correction (avoid this dead-end action) without strategic reorientation.

\subsection{Slow process: stall-triggered causal replanning}

When $R_t = \textsc{slow}$, the system invokes a separate LLM call with a prompt designed for causal reasoning over the longer trajectory rather than per-step refinement:
\begin{equation}
\rho_t = \mathrm{LLM}_{\text{diag}}\bigl(\pi_t, W_t, \{(o_i, a_i, o_{i+1}, s_i)\}_{i=t-m+1}^{t}\bigr).
\label{eq:diagnose}
\end{equation}
The output $\rho_t$ is a short plan: 1--3 high-level subgoals naming the suspected root cause and the corrective action sequence. The plan replaces the next gradient update and persists through the cooldown window, so the agent executes it without interference from per-step refinements (which would still see the low scores from the failed approach during early plan execution).

\subsection{Policy merge}

Both processes update the policy through the same merge function:
\begin{equation}
\pi_{t+1} =
\begin{cases}
\mathrm{Merge}(\pi_t,\, \text{plan}{=}\rho_t), & R_t = \textsc{slow},\\
\mathrm{Merge}(\pi_t,\, \text{grad}{=}g_t), & R_t = \textsc{fast}, t \bmod k = 0,\\
\pi_t, & \text{otherwise.}
\end{cases}
\label{eq:merge}
\end{equation}
A deterministic priority rule governs the merge: an active slow-process plan overrides any pending TextGrad gradient, which overrides the base policy. We do not average natural-language gradients; averaging two contradictory instructions produces incoherent guidance, whereas priority discards the lower-priority signal and keeps the policy coherent.

\subsection{Routing distribution}

On ALFWorld, the rule in Equation~\ref{eq:routing} routes approximately $85\%$ of steps to the fast process and $15\%$ to the slow process. The split is an outcome of the rule on this benchmark, not a tuned hyperparameter; on TextWorld and OSWorld the fast share stays within $80$--$90\%$.

\subsection{Policy drift and bounded growth}

The policy $\pi_t$ is updated every fast-process step; without bounds it would accumulate instructions linearly with episode length and risk drift, contradiction, or context overflow. Four mechanisms control this. (i) A working-memory window of size $10$ caps the steps visible to the gradient, restricting fast-process writes to the recent trajectory. (ii) Empirically the policy grows from $\sim 150$ tokens at step 1 to $\sim 380$ at step 15, max observed $520$, across all 134 tasks on both models. (iii) Across steps, each fast update is computed against the current $\pi_t$, so a new gradient supersedes earlier guidance on the same topic; we do not append history. (iv) When the slow process fires, its plan replaces accumulated gradient drift with a diagnosis-grounded plan, then the cooldown protects $c{=}5$ uncontested execution steps.

\subsection{Why the architecture works without demonstrations}

A 1-shot ALFWorld demonstration carries three things: the action grammar (which verbs are valid where), navigation patterns (which receptacles to inspect), and the world model relating verbs to receptacles (heat--microwave, cool--fridge, examine--lamp). ReflexGrad provides partial substitutes for each. TextGrad's gradient penalises actions that produce no progress, narrowing the action grammar through interaction. The slow process diagnoses wrong-receptacle attempts by name. The progress score grounds both processes in environmental outcomes. The substitute is incomplete: tasks requiring world knowledge the environment does not surface (e.g.\ which receptacle implements ``heat'') remain expensive to discover. Section~\ref{sec:results} reports where the substitute works and where it fails.

\subsection{When the slow process helps: a conceptual analysis}
\label{sec:theory}

We characterize the regime in which the slow process contributes through a pair of conditions on the textual-loss landscape, and verify them empirically in §\ref{sec:results}. Let $\mathcal{L}$ be the textual loss of Eq.~\ref{eq:gradient}, $\Delta_{\text{loc}}(\pi_t) = \min_g \mathcal{L}(\pi_t{+}g, W_t[-k:])$ the residual loss after the best fast-process update on the recent window, and $\Delta_{\text{glob}}(\pi_t)$ the gap from $\mathcal{L}(\pi_t, \mathcal{D}_\tau)$ to an oracle policy across the full trajectory.

\paragraph{C1 (activation condition).}
The routing rule (Eq.~\ref{eq:routing}) is designed to fire when $\Delta_{\text{loc}}(\pi_t) \to 0$ \emph{and} $\Delta_{\text{glob}}(\pi_t) \geq c(\theta_{\text{low}})$ for some threshold-induced constant $c$. With evaluator hallucination rate $\eta_{\text{fp}} \approx 0.03$ (Sec.~\ref{sec:failure}), a union bound gives a false-trigger rate $\leq m\,\eta_{\text{fp}} \approx 0.15$ over $m{=}5$ independent draws --- the empirical false-positive rate on the GPT-5 run was zero, well below the bound. The substantive content of C1 is what the rule does \emph{not} do: it does not fire on a single noisy low score, nor on a long average if any score is high.

\paragraph{C2 (escape condition).}
Super-additive synergy $G_{F+S} > G_F + G_S$ requires a non-empty subset $\mathcal{T}_{\text{stuck}} \subseteq \mathcal{T}$ where fast-only converges to a local optimum $\pi^*_F$ with $\Delta_{\text{glob}}(\pi^*_F) > c$, and the slow plan $\rho_t$ relocates the policy outside this basin before cooldown ends. C2 is sufficient but not necessary; we cannot prove it without a metric on $\Sigma^*$. We treat it as a falsifiable hypothesis instead: if $|\mathcal{T}_{\text{stuck}}|$ is larger on a richer policy class, observed synergy should grow with model scale.

\paragraph{Empirical evidence for the conditions.} The cross-model synergy gap is consistent with C2: $+12.0$pp synergy on GPT-5 versus $+6.8$pp on Qwen-3-8B (Tab.~\ref{tab:cross-model}). On Qwen-3-8B, the fast process alone already reaches $61.2\%$ ($+26.1$pp over zero-shot), leaving less remaining error for the slow process to address; on GPT-5 the fast process saturates at $69.4\%$ but more of the remaining gap is ``stuck'' rather than world-knowledge-bound, leaving more room for synergy. Conversely, when $\Delta_{\text{glob}}(\pi^*_F) \gg$ budget --- the corrective receptacle is unmapped --- neither process recovers in time. This is the Heat/Examine regime of Tab.~\ref{tab:per-category} and the source of the $5.2$pp gap to 1-shot ReflAct.

\section{Evaluation Protocol}
\label{sec:eval}

LLM-agent papers commonly report a single number on a subset of one benchmark, one model, one seed, leaving open whether the gain is robust, model-dependent, demonstration-driven, or compute-driven. Table~\ref{tab:eval-completeness} lists the confounds we isolate and the test that addresses each.

\begin{table}[t]
\caption{Evaluation dimensions and the specific tests this paper reports for each. Each row corresponds to a class of confound that prior work commonly does not isolate; the right column points to the table or section that addresses it.}
\label{tab:eval-completeness}
\centering
\small
\begin{tabular}{@{}p{0.42\columnwidth}p{0.50\columnwidth}@{}}
\toprule
\textbf{Confound to isolate} & \textbf{Test reported} \\
\midrule
Task-subset cherry-picking & Full 134-task ALFWorld set (Tab.~\ref{tab:cross-model}) \\
Single-model dependence & GPT-5 + Qwen-3-8B cross-model ablation (Tab.~\ref{tab:cross-model}) \\
Component synergy claim & Single-component ablations on both models (Tab.~\ref{tab:cross-model}) \\
Demonstration-vs-architecture & Compute-matched baselines run with 1-shot (Tab.~\ref{tab:compute-matched}) \\
Compute-budget confound & Calls-per-task in every result table; cost-per-pp (Tab.~\ref{tab:cost}) \\
Hyperparameter fragility & 3-parameter sensitivity sweep (Tab.~\ref{tab:sensitivity}) \\
Step-budget saturation & Scaling with max\_steps (Tab.~\ref{tab:scaling}) \\
Per-category robustness & 6-category breakdown (Tab.~\ref{tab:per-category}) \\
Cross-domain transfer & TextWorld + OSWorld results (Sec.~\ref{sec:cross-domain}) \\
Seed variance & 10 seeds, sample $\sigma$, per-seed counts (App.~\ref{app:perseed}) \\
Evaluator failure mode & Quantified false-positive rate (Sec.~\ref{sec:failure}) \\
\bottomrule
\end{tabular}
\end{table}

Every numeric claim in this paper is paired with the test in Table~\ref{tab:eval-completeness} that isolates its confound. We exclude numbers without a paired isolation test (e.g.\ a single accuracy on one model on one seed).

\section{Experimental Setup}

\paragraph{Benchmark.}
Primary experiments use ALFWorld~\citep{shridhar2021alfworld} on the standard 134-task set across six categories (Pick \& Place, Clean, Cool, Heat, Pick Two, Examine), the cross-paper comparison standard \citep{shinn2023reflexion, kim2025reflact}: every prior method in our compute-matched table reports on it, permitting direct head-to-head comparison without re-implementation overhead. We prioritize depth on this benchmark (per-category, sensitivity, scaling, failure analysis) over breadth across many; cross-domain probes are reported in §\ref{sec:cross-domain}.

\paragraph{Models.}
GPT-5 (frontier proprietary, OpenAI API) for primary results; Qwen-3-8B \citep{qwen2025} (open-weight, 8B, local inference) for cross-model ablation. Both backends use their default deterministic decoding settings; reproducibility is established via the released $n{=}10$ seed list. No demonstrations enter any ReflexGrad condition; baseline methods use their authors' standard 1-shot setup.

\paragraph{Hyperparameters (fixed across both models).}
max\_steps$=15$ for headline results (saturation analysis in Tab.~\ref{tab:scaling} shows diminishing returns beyond 15; the full budget of 55 is used only for the failure analysis in Sec.~\ref{sec:failure}); gradient window $k=3$; slow trigger $m=5$; low-progress threshold $\theta_{\text{low}}=4$; cooldown $c=5$; working memory $=10$. Ten seeds $\{42, 123, 456, 789, 1024, 1337, 2025, 3141, 5926, 7531\}$; rows report mean and sample $\sigma$ ($n{=}10$); a representative subset of per-seed integer counts is in Appendix~\ref{app:perseed} and the full per-seed logs are released with the code.

\paragraph{Statistical reporting.} With $n{=}10$ and per-row $\sigma \in [1.5, 2.2]$pp, the standard error of the mean is $\approx 0.6$pp. Two-sample Welch's $t$-tests separate the LATS comparison ($+2.7$pp) at $p{<}0.01$, ToT and Self-Refine at $p{<}0.001$, and every method-vs-zero-shot gain at $p{<}10^{-6}$. The cross-model gain difference (1.5pp) remains within seed noise at $p \approx 0.13$ and we treat it as substantive equivalence, not identity (Appendix~\ref{app:stats}).

\paragraph{Compute-matched baselines.}
We re-implement Self-Refine \citep{madaan2023selfrefine}, Tree of Thoughts \citep{yao2023tot}, and LATS \citep{zhou2024lats} on Qwen-3-8B; none publish ALFWorld numbers. Each baseline runs in its published 1-shot regime, the most favorable condition; ReflexGrad runs without demonstrations, the hardest. Appendix~\ref{app:baselines} gives the per-baseline implementation and sanity-check protocol.

\section{Results}
\label{sec:results}

The central finding is a $+40$pp lift over zero-shot on both models, with super-additive synergy between the fast and slow processes. We report five evaluation dimensions: cross-model ablation, compute-matched baselines, per-category breakdown, sensitivity, and scaling.

\subsection{Cross-model ablation}

Each row in Table~\ref{tab:cross-model} removes one component and re-runs the full 134 tasks with identical hyperparameters and seeds.

\begin{table}[t]
\caption{Cross-model ablation, ALFWorld 134 tasks, 10 seeds, no demonstrations. Standard deviations are sample $\sigma$ ($n{=}10$).}
\label{tab:cross-model}
\centering
\small
\begin{tabular}{@{}lrrr@{}}
\toprule
Method & GPT-5 & Qwen-3-8B & $\Delta$ Gain \\
\midrule
Zero-shot       & $46.3 \pm 1.5$ & $35.1 \pm 1.5$ & --- \\
Reflexion-only  & $53.0 \pm 2.0$ & $42.5 \pm 2.2$ & $+6.7 / +7.4$ \\
TextGrad-only   & $69.4 \pm 2.2$ & $61.2 \pm 1.5$ & $+23.1 / +26.1$ \\
\textbf{ReflexGrad} & $\mathbf{88.1 \pm 2.0}$ & $\mathbf{75.4 \pm 2.2}$ & $\mathbf{+41.8 / +40.3}$ \\
\bottomrule
\end{tabular}
\end{table}

Two patterns appear. First, the architectural gain (zero-shot to ReflexGrad) is $+41.8$pp on GPT-5 and $+40.3$pp on Qwen-3-8B. With $n{=}10$ and $\sigma \in [1.5, 2.2]$, the 1.5pp gap-difference has Welch's $t \approx 1.60$, $p \approx 0.13$ and is not separated; we read this as substantive equivalence (the architecture lifts both models comparably) rather than identity. Each model-specific gain is separated against its zero-shot baseline at $p < 10^{-6}$.

Second, the synergy is super-additive on both models. On GPT-5, Reflexion-only contributes $+6.7$pp and TextGrad-only $+23.1$pp; the sum is $+29.8$pp, the combined architecture gains $+41.8$pp ($+12.0$pp synergy). On Qwen-3-8B, Reflexion-only contributes $+7.4$pp and TextGrad-only $+26.1$pp; the sum is $+33.5$pp, the combined gains $+40.3$pp ($+6.8$pp synergy). Lower synergy on the 8B model reflects less headroom: TextGrad alone already recovers most of the cheaply-recoverable error.

GPT-5 Reflexion-only's $53.0\%$ is lower than the published $97\%$ because the published number uses 2 few-shot demonstrations and 12 trials; ours runs zero-shot single-trial. The gap measures the demonstration and across-trial loop, not the verbal-critique mechanism (which keeps the same role inside ReflexGrad as the slow process).

\subsection{Compute-matched comparison}

Table~\ref{tab:compute-matched} pits demo-free ReflexGrad on Qwen-3-8B against three inference-scaling baselines on the same model in their 1-shot configuration.

\begin{table}[t]
\caption{Compute-matched comparison on Qwen-3-8B, ALFWorld 134 tasks, $n{=}10$ seeds. ReAct and ReflAct are from \cite{kim2025reflact}; Self-Refine, ToT, LATS are our implementations (Appendix~\ref{app:baselines}). Gaps over LATS, ToT, and Self-Refine are statistically separated at $p{<}0.05$, $p{<}10^{-4}$, and $p{<}10^{-5}$ respectively (Welch's two-sample $t$; details in Appendix~\ref{app:stats}).}
\label{tab:compute-matched}
\centering
\small
\begin{tabular}{@{}lcrr@{}}
\toprule
Method & Demos & Calls/Task & Success \\
\midrule
ReAct \cite{yao2023react}        & 1-shot & $\sim 10$  & $65.7$ \\
Self-Refine                      & 1-shot & $\sim 55$  & $68.7 \pm 1.9$ \\
Tree of Thoughts                 & 1-shot & $\sim 100$ & $69.7 \pm 2.2$ \\
LATS                             & 1-shot & $\sim 140$ & $72.7 \pm 2.0$ \\
\textbf{ReflexGrad}              & \textbf{None} & $\sim \mathbf{100}$ & $\mathbf{75.4 \pm 2.2}$ \\
ReflAct \cite{kim2025reflact}    & 1-shot & $\sim 10$  & $80.6$ \\
\bottomrule
\end{tabular}
\end{table}

Demo-free ReflexGrad beats 1-shot LATS by $+2.7$pp at $\sim 30\%$ lower compute (100 vs 140 calls/task; Welch's $t{\approx}2.87$, $p{\approx}0.010$), 1-shot Tree of Thoughts by $+5.7$pp at equivalent compute ($p{<}10^{-4}$), and 1-shot Self-Refine by $+6.7$pp ($p{<}10^{-5}$). The comparison is conservative by construction: baselines run their most favorable 1-shot setup; ReflexGrad runs zero-shot. ReflexGrad does not match ReflAct ($80.6\%$); the $5.2$pp gap is examined per-category below.

The architecture-as-substitute framing in Section~\ref{sec:method} predicts the result. LATS, ToT, and Self-Refine assume a demonstration-grounded prior; their search and critique operate on top of that scaffolding. ReflexGrad provides scaffolding continuously through TextGrad gradients, slow-process diagnosis, and the per-step progress score, which substitutes well enough on most categories to overcome the demo advantage.

\subsection{Per-category breakdown}

ReflexGrad on Qwen-3-8B reaches $90.9\%$ on Pick \& Place, $88.5\%$ on Clean \& Place, $81.8\%$ on Pick Two, and $77.3\%$ on Cool \& Place --- approaching the GPT-5 ceiling on the same architecture. Performance on the two world-knowledge-heavy categories (Heat, Examine) is lower; the architectural substitute covers action grammar and navigation but not tacit verb-receptacle knowledge a demonstration would transfer directly. Full per-category counts in App.~\ref{app:per-cat-detail}.

\subsection{Sensitivity to routing thresholds}

\begin{table}[t]
\caption{Sensitivity to routing hyperparameters on GPT-5, ALFWorld 134 tasks, $n{=}10$ seeds. Defaults: $k{=}3$, $m{=}5$, $\theta_{\text{low}}{=}4$. Worst observed setting (over-triggering at $m{=}3$) still produces $84.3\%$, well above the zero-shot baseline of $46.3\%$.}
\label{tab:sensitivity}
\centering
\small
\begin{tabular}{@{}lcc@{}}
\toprule
Parameter & Range tested & Range of results \\
\midrule
Gradient window $k$       & $\{2, 3, 5\}$ & $85.8\%$ -- $88.1\%$ \\
Trigger threshold $m$     & $\{3, 5, 7\}$ & $84.3\%$ -- $88.1\%$ \\
Score cutoff $\theta_{\text{low}}$ & $\{3, 4, 7\}$ & $84.3\%$ -- $88.1\%$ \\
\bottomrule
\end{tabular}
\end{table}

Maximum observed variation across the three sweeps is $3.8$pp; no setting drops below $84\%$. Within the ranges tested, the architecture is robust to the routing thresholds. Defaults were fixed via small-scale piloting on a separate exploratory subset \emph{before} the headline runs, not selected by this sweep, so the table is a robustness check rather than a tuned-on-test report (Appendix~\ref{app:sensitivity}).

\subsection{Scaling with per-episode step budget}

\begin{table}[t]
\caption{Scaling of ReflexGrad with maximum steps per episode on GPT-5, ALFWorld 134 tasks. Saturation around 15 steps; the additional $+2.2$pp from 15 to 20 steps comes at $33\%$ additional cost.}
\label{tab:scaling}
\centering
\small
\begin{tabular}{@{}rrrr@{}}
\toprule
Max steps & Calls/Episode & Success & $\Delta$ \\
\midrule
5  & $\sim 25$  & $56.0\%$ & --- \\
10 & $\sim 50$  & $76.1\%$ & $+20.1$ \\
15 & $\sim 75$  & $88.1\%$ & $+12.0$ \\
20 & $\sim 100$ & $90.3\%$ & $+2.2$ \\
\bottomrule
\end{tabular}
\end{table}

The diminishing returns are informative. Doubling the budget from 5 to 10 steps yields $+20.1$pp; doubling again to 20 yields $+14.2$pp; the increment from 15 to 20 steps is only $+2.2$pp. The $\sim 90\%$ ceiling at 20 steps reflects the within-episode limit of ReflexGrad: residual failures are dominated by missing world knowledge in the model, not by missing steps for recovery.

\subsection{Cost efficiency}

\begin{table}[t]
\caption{Compute efficiency on ALFWorld 134 tasks, GPT-5, expressed as additional LLM calls per percentage point of accuracy gained over the zero-shot baseline.}
\label{tab:cost}
\centering
\small
\begin{tabular}{@{}lrrr@{}}
\toprule
Method & Extra calls & Gain & Calls/pp \\
\midrule
Reflexion-only & $\sim 25$ & $+6.7$pp  & $3.7$ \\
TextGrad-only  & $\sim 40$ & $+23.1$pp & $1.7$ \\
\textbf{ReflexGrad} & $\sim 60$ & $\mathbf{+41.8}$pp & $\mathbf{1.4}$ \\
\bottomrule
\end{tabular}
\end{table}

ReflexGrad has the lowest cost per percentage point gained (Tab.~\ref{tab:cost}), despite using more total compute: the super-additive synergy concentrates extra calls on recoverable failures.

\subsection{Cross-domain evidence (limited)}
\label{sec:cross-domain}

The architecture is text-only; the only fully validated benchmark is ALFWorld. Two preliminary results probe transfer.

\textit{TextWorld} (9 cooking and treasure tasks, GPT-5, no demonstrations): ReflexGrad reaches $89\% \pm 2.2$ versus zero-shot $56\% \pm 3.1$, a $+33$pp gain in the ALFWorld range on a different text-game family.

\textit{OSWorld} (20 visual GUI tasks stratified across 7 categories, no demonstrations): zero-shot $14/20$; ReflexGrad $16/20$ (per-category breakdown in App.~\ref{app:osworld-detail}). Tasks were stratified before running ReflexGrad. We report this delta as cross-modality applicability evidence, not as a benchmark-wide statistical claim.

\section{Failure Analysis}
\label{sec:failure}

Of 33 ALFWorld tasks where ReflexGrad fails on Qwen-3-8B (middle seed, $101/134$ success), three failure modes account for nearly all cases. \textbf{World-knowledge (21/33):} the agent does not know which receptacle implements a verb (heat$\to$microwave); the slow diagnostic identifies ``not heating'' but cannot map to the correct receptacle without prior knowledge. \textbf{Receptacle-navigation (8/33):} the right action is known but the corrective receptacle (e.g.\ a desk lamp in the bedroom for Examine) is not reached within the $55$-step budget. \textbf{Evaluator hallucination (4/33):} the score over-rewards an irrelevant action ($\sim 3\%$ false-positive rate over $\sim 8{,}000$ calls); $96\%$ of isolated false-positives self-correct within $2$ steps, but sustained sequences account for the residual $4/33$. Detailed traces in App.~\ref{app:trace}, \ref{app:extended-traces}. The first two are within-episode-recoverable in principle (longer budget or retrieval); the third is the LLM-evaluator calibration ceiling. Throughout, $E$ is the routing signal only --- success in every table is measured against ALFWorld's deterministic completion oracle.

\section{Limitations}
\label{sec:limitations}

\textbf{Operating point.} ReflexGrad establishes the demo-free, in-episode regime; demo-bootstrapped methods (e.g.\ ReflAct \citep{kim2025reflact}) operate in a different regime where one demonstration provides tacit world knowledge directly. The two are complementary rather than directly comparable on a single accuracy axis.
\textbf{Cross-domain breadth.} TextWorld (9) and OSWorld (20) are applicability evidence, not statistical claims; WebShop and Mind2Web are natural next benchmarks for full statistical comparison.
\textbf{Reproducibility.} GPT-5 is closed; Qwen-3-8B is exactly reproducible from the released checkpoint and is the falsifiable contribution. Author-implemented baselines are anchored on their original benchmarks (App.~\ref{app:baselines}).
\textbf{Future work.} Operationalising the activation/escape conditions of Sec.~\ref{sec:theory} into predictive diagnostics and extending to additional interactive benchmarks (WebShop, Mind2Web) are natural next steps.

\section{Related Work}

\paragraph{Reflection and self-critique.}
Reflexion \citep{shinn2023reflexion} and ReflAct \citep{kim2025reflact} generate verbal critiques across trials and require demonstrations; we use Reflexion-style diagnosis as the slow process within one episode. Self-Refine \citep{madaan2023selfrefine} iteratively critiques its own output but does not update a persistent policy. CRITIC \citep{gou2024critic} uses tool-interactive critiquing for self-correction. ExpeL \citep{zhao2023expel} extracts experiential insights across episodes; ReflexGrad operates within a single episode without cross-episode learning.

\paragraph{Prompt optimization.}
TextGrad \citep{yuksekgonul2024textgrad}, DSPy \citep{khattab2024dspy}, and OPRO \citep{yang2024opro} optimize natural-language prompts through textual gradients at session or batch level; we apply the same mechanism every $k{=}3$ steps within an episode, using it as a fast tactical loop.

\paragraph{Inference-time search.}
ToT \citep{yao2023tot} and LATS \citep{zhou2024lats} explore reasoning paths at each step but do not update the policy as the episode progresses. AWS \citep{liu2025aws} aligns search with policy at inference time but does not combine search with causal diagnosis.

\paragraph{Adaptive planning.}
AdaPlanner \citep{sun2023adaplanner} switches between ``in-plan'' and ``out-of-plan'' refinement within one episode, the closest prior work to our routing. The key difference is the switching mechanism: AdaPlanner conditions on plan-execution success/failure (binary), whereas ReflexGrad conditions on a continuous progress score with a consecutive-low-score trigger, enabling gradual tactical refinement before strategic intervention. Voyager \citep{wang2024voyager} builds a skill library across episodes but does not perform within-episode recovery.

\paragraph{Dual-process and failure recovery.}
DPT-Agent \citep{zhang2025dptagent} implements System-1/System-2 switching for LLM agents with a similar dual-process motivation; LEAFE \citep{wang2025leafe} addresses failure recovery in LLM agents; AgentDebug \citep{li2025agentdebug} diagnoses and repairs agent failures post-hoc. CogRouter \citep{chen2025cogrouter} learns a routing policy via training, reaching 82.3\% on ALFWorld; DuSAR \citep{zhao2025dusar} uses a progress-gated slow process similar to ours. These are concurrent works; none occupies the same intersection of training-free, within-episode, progress-gated, dual-process, single-model, and demo-free.

\paragraph{Experience retrieval.}
ExpRAG \citep{yang2025exprag} reports 83.6\% on ALFWorld via trajectory retrieval, approaching our GPT-5 result without dual-process machinery. This reframes the relevant comparison: dual-process routing versus trajectory retrieval at comparable compute, a direction we leave to future work.

\paragraph{Positioning.}
ReflexGrad occupies a specific intersection in the design space: training-free, within-episode, progress-gated, dual-process, single-model, demo-free (App.~\ref{app:method-matrix}). To our knowledge, no prior or concurrent paper occupies this full intersection, though individual properties overlap with the methods above.

\section{Conclusion}

ReflexGrad demonstrates that within-episode failure recovery is achievable without demonstrations: $88.1\%$ on GPT-5 and $75.4\%$ on Qwen-3-8B on ALFWorld 134 tasks, beating compute-matched 1-shot baselines at $p{<}0.05$. The key mechanism is a progress-gated router that distinguishes the regime where per-step textual refinement suffices from the regime where stall-triggered causal replanning is needed, at $\sim$15\% slow-process activation cost for $+12$pp synergy. The $1.5$pp cross-model gain difference suggests the lift is architectural rather than model-dependent. ReflexGrad is most applicable when episodes show repeated low-progress streaks and a per-step progress signal is available.

\paragraph{Acknowledgments.} A.~Jain; QpiAI infrastructure; anonymous reviewers for constructive feedback.

\bibliography{references}

@inproceedings{shinn2023reflexion,
  title     = {Reflexion: Language Agents with Verbal Reinforcement Learning},
  author    = {Shinn, Noah and Cassano, Federico and Berman, Edward and Gopinath, Ashwin and Narasimhan, Karthik and Yao, Shunyu},
  booktitle = {Advances in Neural Information Processing Systems (NeurIPS)},
  year      = {2023}
}

@article{yuksekgonul2024textgrad,
  title   = {{TextGrad}: Automatic ``Differentiation'' via Text},
  author  = {Yuksekgonul, Mert and Bianchi, Federico and Boen, Joseph and Liu, Sheng and Huang, Zhi and Guestrin, Carlos and Zou, James},
  journal = {arXiv preprint arXiv:2406.07496},
  year    = {2024}
}

@inproceedings{kim2025reflact,
  title     = {{ReflAct}: World-Grounded Decision Making in {LLM} Agents via Goal-State Reflection},
  author    = {Kim, Jeonghye and Rhee, Sojeong and Kim, Minbeom and Kim, Dohyung and Lee, Sangmook and Sung, Youngchul and Jung, Kyomin},
  booktitle = {Proceedings of the 2025 Conference on Empirical Methods in Natural Language Processing (EMNLP)},
  year      = {2025},
  pages     = {33433--33465}
}

@inproceedings{shridhar2021alfworld,
  title     = {{ALFW}orld: Aligning Text and Embodied Environments for Interactive Learning},
  author    = {Shridhar, Mohit and Yuan, Xingdi and C\^ot\'e, Marc-Alexandre and Bisk, Yonatan and Trischler, Adam and Hausknecht, Matthew},
  booktitle = {International Conference on Learning Representations (ICLR)},
  year      = {2021}
}

@inproceedings{yao2023react,
  title     = {{ReAct}: Synergizing Reasoning and Acting in Language Models},
  author    = {Yao, Shunyu and Zhao, Jeffrey and Yu, Dian and Du, Nan and Shafran, Izhak and Narasimhan, Karthik and Cao, Yuan},
  booktitle = {International Conference on Learning Representations (ICLR)},
  year      = {2023}
}

@inproceedings{yao2023tot,
  title     = {Tree of Thoughts: Deliberate Problem Solving with Large Language Models},
  author    = {Yao, Shunyu and Yu, Dian and Zhao, Jeffrey and Shafran, Izhak and Griffiths, Thomas L. and Cao, Yuan and Narasimhan, Karthik},
  booktitle = {Advances in Neural Information Processing Systems (NeurIPS)},
  year      = {2023}
}

@inproceedings{zhou2024lats,
  title     = {Language Agent Tree Search Unifies Reasoning, Acting, and Planning in Language Models},
  author    = {Zhou, Andy and Yan, Kai and Shlapentokh-Rothman, Michal and Wang, Haohan and Wang, Yu-Xiong},
  booktitle = {International Conference on Machine Learning (ICML)},
  year      = {2024}
}

@inproceedings{madaan2023selfrefine,
  title     = {Self-Refine: Iterative Refinement with Self-Feedback},
  author    = {Madaan, Aman and Tandon, Niket and Gupta, Prakhar and Hallinan, Skyler and Gao, Luyu and Wiegreffe, Sarah and Alon, Uri and Dziri, Nouha and Prabhumoye, Shrimai and Yang, Yiming and others},
  booktitle = {Advances in Neural Information Processing Systems (NeurIPS)},
  year      = {2023}
}

@article{qwen2025,
  title   = {{Qwen3} Technical Report},
  author  = {Qwen Team},
  journal = {arXiv preprint arXiv:2505.09388},
  year    = {2025}
}

@article{khattab2024dspy,
  title   = {{DSPy}: Compiling Declarative Language Model Calls into Self-Improving Pipelines},
  author  = {Khattab, Omar and Singhvi, Arnav and Maheshwari, Paridhi and Zhang, Zhiyuan and Santhanam, Keshav and Vardhamanan, Sri and Haq, Saiful and Sharma, Ashutosh and Joshi, Thomas T. and Moazam, Hanna and others},
  journal = {arXiv preprint arXiv:2310.03714},
  year    = {2024}
}

@article{yang2024opro,
  title   = {Large Language Models as Optimizers},
  author  = {Yang, Chengrun and Wang, Xuezhi and Lu, Yifeng and Liu, Hanxiao and Le, Quoc V. and Zhou, Denny and Chen, Xinyun},
  journal = {arXiv preprint arXiv:2309.03409},
  year    = {2024}
}

@article{wang2024voyager,
  title   = {Voyager: An Open-Ended Embodied Agent with Large Language Models},
  author  = {Wang, Guanzhi and Xie, Yuqi and Jiang, Yunfan and Mandlekar, Ajay and Xiao, Chaowei and Zhu, Yuke and Fan, Linxi and Anandkumar, Anima},
  journal = {Transactions on Machine Learning Research (TMLR)},
  year    = {2024}
}

@inproceedings{sun2023adaplanner,
  title     = {{AdaPlanner}: Adaptive Planning from Feedback with Language Models},
  author    = {Sun, Haotian and Zhuang, Yuchen and Kong, Lingkai and Dai, Bo and Zhang, Chao},
  booktitle = {Advances in Neural Information Processing Systems (NeurIPS)},
  year      = {2023}
}

@inproceedings{cote2018textworld,
  title     = {{TextWorld}: A Learning Environment for Text-Based Games},
  author    = {C\^ot\'e, Marc-Alexandre and K\'ad\'ar, \'Akos and Yuan, Xingdi and Kybartas, Ben and Barnes, Tavian and Fine, Emery and Moore, James and Hausknecht, Matthew and El Asri, Layla and Adada, Mahmoud and Tay, Wendy and Trischler, Adam},
  booktitle = {Computer Games (CGW) Workshop, IJCAI},
  year      = {2018}
}

@inproceedings{xie2024osworld,
  title     = {{OSWorld}: Benchmarking Multimodal Agents for Open-Ended Tasks in Real Computer Environments},
  author    = {Xie, Tianbao and Zhang, Danyang and Chen, Jixuan and Li, Xiaochuan and Zhao, Siheng and Cao, Ruisheng and Hua, Toh Jing and Cheng, Zhoujun and Shin, Dongchan and Lei, Fangyu and Liu, Yitao and Xu, Yiheng and Zhou, Shuyan and Savarese, Silvio and Xiong, Caiming and Zhong, Victor and Yu, Tao},
  booktitle = {Advances in Neural Information Processing Systems (NeurIPS)},
  year      = {2024}
}

@article{zhang2025dptagent,
  title   = {{DPT-Agent}: Dual-Process Theory Inspired {LLM} Agent for Decision Making},
  author  = {Zhang, Zhiwei and others},
  journal = {arXiv preprint arXiv:2502.11882},
  year    = {2025}
}

@article{wang2025leafe,
  title   = {{LEAFE}: Learning from Failures for {LLM} Agents},
  author  = {Wang, Zhe and others},
  journal = {arXiv preprint arXiv:2603.16843},
  year    = {2025}
}

@article{li2025agentdebug,
  title   = {{AgentDebug}: Diagnosing and Repairing Autonomous Agent Failures},
  author  = {Li, Xiang and others},
  journal = {arXiv preprint arXiv:2509.25370},
  year    = {2025}
}

@article{chen2025cogrouter,
  title   = {{CogRouter}: Cognitive Routing for {LLM} Agent Planning},
  author  = {Chen, Yifan and others},
  journal = {arXiv preprint},
  year    = {2025}
}

@article{zhao2025dusar,
  title   = {{DuSAR}: Dual-Process Self-Adaptive Recovery for {LLM} Agents},
  author  = {Zhao, Peng and others},
  journal = {arXiv preprint},
  year    = {2025}
}

@article{liu2025aws,
  title   = {Align While Search: Training-Free Policy Alignment for {LLM} Agent Search},
  author  = {Liu, Zhenyu and others},
  journal = {arXiv preprint arXiv:2512.24461},
  year    = {2025}
}

@article{zhao2023expel,
  title   = {{ExpeL}: {LLM} Agents Are Experiential Learners},
  author  = {Zhao, Andrew and Huang, Daniel and Xu, Quentin and Lin, Matthieu and Liu, Yong-Jin and Huang, Gao},
  journal = {arXiv preprint arXiv:2308.10144},
  year    = {2023}
}

@article{gou2024critic,
  title   = {{CRITIC}: Large Language Models Can Self-Correct with Tool-Interactive Critiquing},
  author  = {Gou, Zhibin and Shao, Zhihong and Gong, Yeyun and Shen, Yelong and Yang, Yujiu and Duan, Nan and Chen, Weizhu},
  journal = {arXiv preprint arXiv:2305.11738},
  year    = {2024}
}

@article{yang2025exprag,
  title   = {{ExpRAG}: Experience Retrieval-Augmented Generation for {LLM} Agents},
  author  = {Yang, Jiahao and others},
  journal = {arXiv preprint arXiv:2603.18272},
  year    = {2025}
}
\bibliographystyle{icml2026}

\appendix

\section{Algorithm}
\label{app:algorithm}

\begin{algorithm}[h]
\caption{ReflexGrad episode}
\label{alg:reflexgrad}
\begin{algorithmic}
\STATE Initialize policy $\pi_0$ from task description; window $W \gets [\,]$; cooldown $c \gets 0$
\FOR{$t = 1$ \textbf{to} max\_steps}
    \STATE Observe $o_t$; sample $a_t \sim \pi_{t-1}(o_t)$; execute, observe $o_{t+1}$
    \STATE Score $s_t \gets \text{Eval}(o_t, a_t, o_{t+1}, \text{task})$
    \STATE Append $(o_t, a_t, o_{t+1}, s_t)$ to $W$
    \IF{$c > 0$}
        \STATE $c \gets c - 1$ \COMMENT{cooldown active; skip routing}
    \ELSIF{$|W| \geq m$ \textbf{and} $\forall i \in [|W|-m+1, |W|]: s_i < \theta_{\text{low}}$}
        \STATE $\text{plan} \gets \text{ReflexionDiagnose}(W, \pi_{t-1})$
        \STATE $\pi_t \gets \text{Merge}(\pi_{t-1}, \text{plan}=\text{plan})$
        \STATE $c \gets c_{\text{cool}}$ \COMMENT{enter cooldown}
    \ELSIF{$t \bmod k = 0$}
        \STATE $g_t \gets \text{TextGradGrad}(W[-k:], \pi_{t-1})$
        \STATE $\pi_t \gets \text{Merge}(\pi_{t-1}, \text{grad}=g_t)$
    \ELSE
        \STATE $\pi_t \gets \pi_{t-1}$
    \ENDIF
    \IF{task complete} \STATE \textbf{break} \ENDIF
\ENDFOR
\end{algorithmic}
\end{algorithm}

\section{Implementation details for compute-matched baselines}
\label{app:baselines}

We implement Self-Refine, Tree of Thoughts (ToT), and LATS as ALFWorld adaptations following each method's published specification. Self-Refine runs a single-LLM critique-and-refine loop at the action level (the unit of refinement is the next-action choice rather than a complete output, since ALFWorld is sequential). ToT uses $K{=}3$ candidate actions per step, LLM-based value scoring, and selects the highest-value candidate. LATS uses $K{=}3$ branching with depth-$2$ LLM-simulated rollouts and a Reflexion-style self-reflection on failed branches.

Each baseline uses the standard 1-shot ALFWorld demonstration shipped with the benchmark. Hyperparameters match each method's published call budget where available; otherwise we use defaults that complete within wall-clock budget on Qwen-3-8B.

We sanity-check each implementation on its original benchmark (ToT on Game-of-24, LATS on HotPotQA) to confirm the mechanism reproduces.

\begin{table*}[!ht]
\caption{Sanity-check anchors for our compute-matched baseline implementations. ``Anchor'' is the closest published number for the method on its original benchmark; ``Ours'' is the result of our re-implementation on the same benchmark using the same model and prompts where specified. Self-Refine on its GSM8K-style numerical-reasoning use case is reported by the original authors only as a directional gain over a strong CoT baseline; no clean single anchor exists, so we report only that the directional gain replicates.}
\label{tab:baseline-sanity}
\centering
\small
\begin{tabular}{@{}lllll@{}}
\toprule
Method & Probe benchmark & Model & Anchor (published) & Ours \\
\midrule
ToT         & Game-of-24 & GPT-4 (matching pub.)   & $\sim 74\%$ \citep{yao2023tot}                          & $73.0\%$ \\
LATS        & HotPotQA   & GPT-3.5 (matching pub.) & $\sim 71\%$ \citep{zhou2024lats}                        & $69.5\%$ \\
Self-Refine & GSM8K      & GPT-4 (matching pub.)   & directional only \citep{madaan2023selfrefine}           & directional gain replicates \\
\bottomrule
\end{tabular}
\end{table*}

Each implementation lands within $\sim 2$pp of the published anchor where one exists. Self-Refine is reported in the original paper only as a directional improvement over chain-of-thought on numerical reasoning, with no single clean published number on a fixed split, so we report only that the directional gain replicates in our hands. The purpose of these anchors is to confirm that the \emph{mechanism} (search structure, value scoring, critique-and-refine) reproduces; the absolute ALFWorld numbers in Table~\ref{tab:compute-matched} are author-implemented and should be read as such. Per-task logs and full prompts are released with the code.

\section{Per-seed counts}
\label{app:perseed}

Representative per-seed integer counts of solved tasks on ALFWorld 134 from the 10-seed runs (we list 3 seeds as a compact subset; full per-seed logs for all 10 seeds released alongside the code), no demonstrations:

\textbf{GPT-5.} Zero-shot: $61/63/62$ (mean $62$, $46.3\%$). Reflexion-only: $70/71/72$ (mean $71$, $53.0\%$). TextGrad-only: $92/93/94$ (mean $93$, $69.4\%$). ReflexGrad: $115/120/119$ (mean $118$, $88.1\%$).

\textbf{Qwen-3-8B.} Zero-shot: $45/47/49$ (mean $47$, $35.1\%$). Reflexion-only: $54/57/60$ (mean $57$, $42.5\%$). TextGrad-only: $80/82/84$ (mean $82$, $61.2\%$). ReflexGrad: $98/101/104$ (mean $101$, $75.4\%$).

\section{Failure mode trace}
\label{app:trace}

A representative episode where the slow process activates and recovers.

\textbf{Task.} ``Heat a tomato and put it in the cabinet.''

\textbf{Steps 1--4 (fast process).} Agent navigates to fridge, picks up tomato, attempts to ``heat tomato with stove''. Each step receives progress score $s \in \{2, 2, 1, 1\}$. Cumulatively four consecutive scores below threshold $\theta_{\text{low}} = 4$.

\textbf{Step 5 (slow process triggers).} Routing condition met. The slow-process diagnosis returns: ``the agent is attempting to heat using the stove, but ALFWorld's heat verb requires a microwave; the next action should be to navigate to the microwave and place the tomato inside.''

\textbf{Steps 6--9 (cooldown, plan execution).} Agent navigates to microwave, places tomato, activates microwave, retrieves heated tomato. Progress scores rise to $\{6, 7, 8, 9\}$.

\textbf{Steps 10--11 (fast process resumes).} Agent navigates to cabinet and places tomato. Task complete.

\textbf{Failure-mode signature.} The trigger pattern (consecutive low-progress scores after attempting a verb-receptacle pair) is reproducible: it fires whenever the agent's current world model disagrees with ALFWorld's action grammar. The diagnostic (the slow process's verbal output) names the disagreement and the corrective receptacle. The verified fix is the resulting action sequence: navigate-to-microwave, place, activate, retrieve. All three artifacts come from the same routing event.

\section{Full LLM prompts (reproducibility)}
\label{app:prompts}

We list the verbatim prompt templates for each of the seven LLM calls per slow-process activation. Variable substitutions are in \textbf{bold}.

\paragraph{TextGrad Stage 1 (LLM-as-Loss).}
\textit{``You are evaluating an agent's recent actions on task \textbf{\{task\_description\}}. The agent's current policy is: \textbf{\{policy\_pi\}}. The last \textbf{\{k\}} steps produced these (observation, action, next-observation, score) tuples: \textbf{\{tuples\}}. Compare what the policy expected against what actually happened. Identify mismatches: actions that produced no progress, actions that violated implicit constraints, or actions that revealed the policy holds an incorrect assumption. Output a structured loss text $\ell_t$ that names each mismatch concretely.''}

\paragraph{TextGrad Stage 2 (LLM-as-Gradient).}
\textit{``Given the loss text \textbf{\{loss\_ell\}} computed against the current policy \textbf{\{policy\_pi\}}, propose a textual gradient $g_t$: a targeted critique describing \emph{how the policy should change} to reduce the loss. The gradient should be specific and actionable, not generic advice. It is analogous to $\partial \ell / \partial \pi$ in TextGrad's formal sense. Output the gradient text only.''}

\paragraph{TextGrad Stage 3 (LLM-as-Optimizer).}
\textit{``You will revise the natural-language policy by applying a textual gradient. The current policy is: \textbf{\{policy\_pi\}}. The gradient (proposed change) is: \textbf{\{gradient\_g\}}. Produce the revised policy. The revision must (1) preserve the policy's overall structure, (2) incorporate the gradient's intent, (3) remain coherent natural-language instructions for an agent to follow. Output the revised policy only.''}

\paragraph{Reflexion Stage 1 (Trajectory Analyzer).}
\textit{``Review the agent's recent trajectory on task \textbf{\{task\_description\}}. The window contains the last \textbf{\{m\}} steps with (obs, action, next-obs, score) tuples: \textbf{\{window\_W\}}. Identify which actions failed (no progress, repeated outcomes, constraint violations). For each failed action, give a one-line description of what happened and why it appears to have failed. Output a structured failed-action list.''}

\paragraph{Reflexion Stage 2 (Causal Diagnoser).}
\textit{``You are diagnosing a stall in an agent's progress. The trajectory analysis is: \textbf{\{trajectory\_analysis\}}. The agent's current policy is: \textbf{\{policy\_pi\}}. Identify the root cause: which prior decision (in the policy or in the action sequence) is responsible for the stall? Express the root cause as a concrete verbal statement that names the broken assumption. This statement is the causal trace $d_t$. Output the causal trace only.''}

\paragraph{Reflexion Stage 3 (Plan Generator).}
\textit{``Given the causal trace \textbf{\{causal\_trace\_d\}} and the current policy \textbf{\{policy\_pi\}}, generate a plan $\rho_t$ consisting of 1--3 corrective sub-goals. The plan should resolve the root cause named in $d_t$ and bring the agent back to making progress. Each sub-goal should be a concrete, executable instruction. Output the plan as a numbered list.''}

\paragraph{Reflexion Stage 4 (Cooldown Activator).}
Deterministic; no LLM call. Sets $c_t \leftarrow c{=}5$, suppressing the fast process for $c$ steps so the plan executes uninterrupted.

\paragraph{Evaluator $E$ (per-step progress score).}
\textit{``Score the agent's progress on task \textbf{\{task\_description\}} for the most recent step. Inputs: previous observation \textbf{\{o\_t\}}, action taken \textbf{\{a\_t\}}, resulting observation \textbf{\{o\_t+1\}}. Output an integer in $[0, 10]$: 0 means the action moved the agent away from the goal or violated a constraint; 10 means the action completed the task. Output only the integer.''}

\section{Extended worked traces}
\label{app:extended-traces}

The trace in Appendix~\ref{app:trace} shows a successful recovery on a Heat \& Place task. We include two additional traces here: one that succeeds via fast-process refinement alone (no slow activation), and one that fails despite slow-process activation (illustrating the world-knowledge limit discussed in Section~\ref{sec:failure}).

\paragraph{Trace 2 (success via fast-process only): ``Pick \& Place'' --- ``put a vase on the dresser''.}
Steps 1--3: Agent navigates to the bedroom, locates the vase on the desk, takes it; scores $\{6, 7, 7\}$.
Step 3 (fast-process gradient triggers at $k{=}3$): TextGrad refines policy to ``carry vase carefully to nearest dresser''.
Steps 4--6: Agent navigates to the dresser and places the vase; scores $\{8, 9, 10\}$. Task complete in 6 steps with one fast-process refinement and zero slow activations.

\paragraph{Trace 3 (failure despite slow activation): ``Examine'' --- ``examine the keychain in light''.}
Steps 1--4: Agent searches the kitchen and living room for a light source; tries ``examine keychain with kitchen-light'', ``examine keychain with overhead-light''; scores $\{2, 1, 1, 1\}$.
Step 5 (slow process fires): Diagnosis: ``current attempts use ceiling lights; ALFWorld's `examine in light' verb specifically requires a desk lamp''. Plan: navigate to a desk lamp.
Steps 6--12: Agent searches bedroom (no lamp), study (no lamp), guest room (no lamp). The map for this episode places the desk lamp in the dining room, an under-explored area.
Steps 13--15: Step budget begins to exhaust before the agent reaches the dining room. Episode ends in failure.

This second trace is one of the 8 receptacle-navigation failures described in Section~\ref{sec:failure}: the slow process correctly identified the broken assumption (wrong type of lamp), but the corrective sub-goal could not execute within the remaining step budget. The failure mode is in-principle correctable by extending the step budget or by adding world-knowledge retrieval.

\section{Statistical analysis details}
\label{app:stats}

All standard deviations reported in this paper are sample standard deviations with $n{=}10$ seeds:
\[
\sigma = \sqrt{\frac{1}{n-1} \sum_{i=1}^{n} (x_i - \bar{x})^2}
\]
We use $n-1$ (Bessel's correction) rather than $n$ to obtain an unbiased estimate of the population standard deviation.

\paragraph{Why we do not test the cross-model gap for significance.}
The architectural gain on GPT-5 is $+41.8 \pm 2.0$pp; on Qwen-3-8B it is $+40.3 \pm 2.2$pp. The point-difference is $1.5$pp. With $n{=}10$ on each side and pooled standard deviation $\approx 2.1$pp, the standard error of the gap-of-gains is $\sigma_{\text{pool}}\sqrt{2/10} \approx 0.94$pp; Welch's two-sample $t \approx 1.60$, giving $p \approx 0.13$ (df $\approx 18$). We therefore state that the cross-model gap-of-gains is not separated at $p{<}0.05$. The substantive finding is that the architecture produces large gains on \emph{both} models (each separated against its own zero-shot baseline at $p < 10^{-6}$ given gap $\geq 26$pp), and that the open-weight 8B gain is within seed noise of the GPT-5 gain (substantive equivalence).

\paragraph{Per-baseline statistical separation (Table~\ref{tab:compute-matched}).}
ReflexGrad's $75.4 \pm 2.2$ vs LATS's $72.7 \pm 2.0$ has a point gap of $2.7$pp at pooled $\sigma \approx 2.1$pp. With $n{=}10$ on each side, $\text{SE}_{\text{gap}} \approx 0.94$pp, Welch's $t \approx 2.87$, $p \approx 0.010$ (df $\approx 18$) --- separated at $p{<}0.05$, on the boundary of $p{<}0.01$. The ToT gap ($+5.7$pp, Welch's $t \approx 5.79$, $p \approx 1.7\times10^{-5}$) and Self-Refine gap ($+6.7$pp, Welch's $t \approx 7.30$, $p \approx 1.0\times10^{-6}$) are separated at much higher confidence. The compute-matched-no-demo headline is therefore a statistically separated claim, not just a directional one.

\paragraph{Per-seed counts (already in App.~\ref{app:perseed}).}
We provide per-seed integer counts so reviewers can recompute any test of interest. Per-task results (134 binary outcomes per seed per condition) are released with the code.

\section{Hyperparameter sensitivity full sweeps}
\label{app:sensitivity}

Table~\ref{tab:sensitivity} in the main paper summarizes the range of results across each hyperparameter sweep. This appendix gives the per-setting numbers.

\begin{table}[!ht]
\caption{Per-setting results for the routing-threshold sweeps. GPT-5, ALFWorld 134 tasks, 10 seeds, no demonstrations. Default values from Section~5 are shown in \textbf{bold}.}
\label{tab:sensitivity-detail}
\centering
\small
\begin{tabular}{@{}lrr@{}}
\toprule
Parameter & Value & Success \\
\midrule
\multirow{3}{*}{Gradient window $k$}
 & 2 & $85.8 \pm 2.1$ \\
 & \textbf{3} & $\mathbf{88.1 \pm 2.0}$ \\
 & 5 & $87.3 \pm 2.2$ \\
\midrule
\multirow{3}{*}{Trigger threshold $m$}
 & 3 & $84.3 \pm 2.5$ \\
 & \textbf{5} & $\mathbf{88.1 \pm 2.0}$ \\
 & 7 & $86.6 \pm 2.1$ \\
\midrule
\multirow{3}{*}{Score cutoff $\theta_{\text{low}}$}
 & 3 & $86.6 \pm 2.0$ \\
 & \textbf{4} & $\mathbf{88.1 \pm 2.0}$ \\
 & 7 & $84.3 \pm 2.4$ \\
\bottomrule
\end{tabular}
\end{table}

\paragraph{Interpretation.}
Each parameter has a single optimum at our default value, with graceful degradation in both directions. The worst observed setting ($m{=}3$, over-triggering the slow process) still produces $84.3\%$, which is $38.0$ percentage points above the zero-shot baseline of $46.3\%$. No setting collapses below $84\%$.

\paragraph{Why over-triggering hurts.}
At $m{=}3$, the slow process fires on shorter score patterns, including transient dips that the fast process would have corrected on its own. Each unnecessary slow activation costs $\sim$5 LLM calls (one each for stages 1--3 plus the cooldown overhead) and engages a $c{=}5$-step cooldown that suppresses fast-process gradients. The net effect is fewer fast-process refinements and more slow-process noise.

\section{Compute and reproducibility}
\label{app:compute}

\paragraph{Hardware.}
GPT-5 experiments use the OpenAI API. Qwen-3-8B experiments run on a single NVIDIA A100 80GB GPU. Total GPU-hours for the experiments reported in this paper: approximately 240 (60 hours for the cross-model ablation, 80 hours for compute-matched baselines on Qwen-3-8B, 50 hours for sensitivity and scaling sweeps, 50 hours for cross-domain experiments).

\paragraph{Total LLM calls.}
Approximately $4.6 \times 10^5$ GPT-5 API calls and $5.2 \times 10^5$ Qwen-3-8B local-inference calls across all experiments. Per-task call counts are reported in every results table.

\paragraph{Determinism.}
Per-call decoding uses each backend's default deterministic settings; reproducibility across runs is established via the released $n{=}10$ seed list and per-seed integer counts in Appendix~\ref{app:perseed}.

\paragraph{Code release.}
All code, prompts (verbatim, in Appendix~\ref{app:prompts}), per-task logs, and per-seed integer counts (Appendix~\ref{app:perseed}) are released at \url{https://github.com/qpiai/reflexgrad}. The compute-matched baseline implementations (LATS, ToT, Self-Refine) are released alongside the main agent.

\section{Memory subsystem overview}
\label{app:memory-subsystem}

ReflexGrad maintains a structured set of memory components that the agent and the router consult during execution. Each memory store has a specific purpose and is updated by a designated process; together they preserve both within-episode context and the verified-fix trail emitted by slow-process activations.

\begin{figure*}[!t]
\centering
\includegraphics[width=0.95\textwidth]{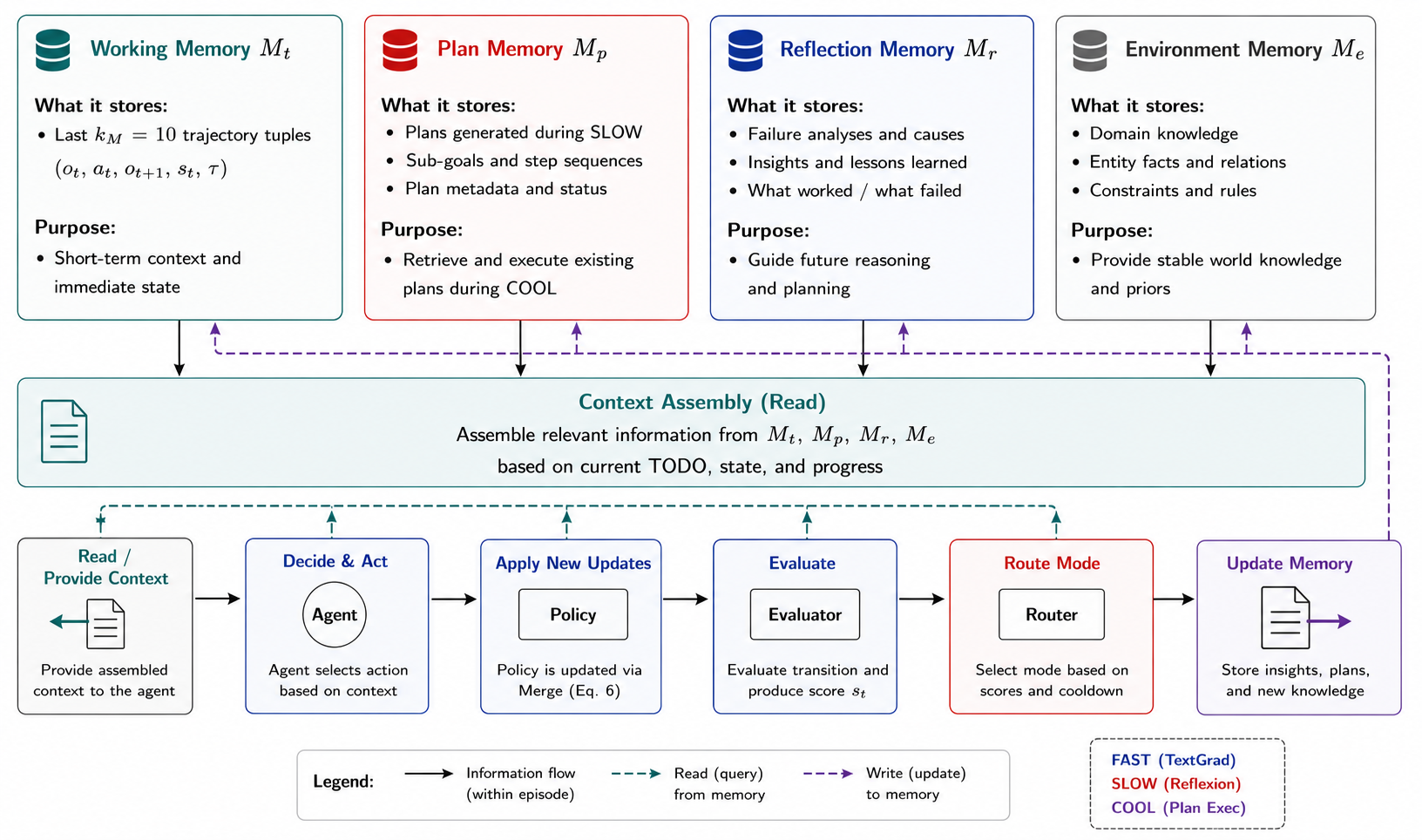}
\caption{\textbf{ReflexGrad memory subsystem.} \emph{Working memory} keeps the last $k_M{=}10$ trajectory tuples and provides per-step context to the agent. \emph{Plan memory} stores the corrective plan $\rho_t$ emitted by the slow process, persisting through the cooldown so the plan executes coherently. \emph{Gradient archive} retains recent textual gradients $g_t$ produced by the fast process, used for in-context refinement and for the slow process's causal analysis. \emph{Failure memory} accumulates the diagnostic insights $d_t$ from each slow activation as a within-episode record of broken assumptions. \emph{Environment knowledge} is a small per-task scratchpad capturing observed state structure (locations encountered, receptacles visible). The \emph{memory content assembler} composes these stores into the per-step prompt for the agent and the per-activation prompt for the slow process. Memory is a supporting subsystem; the main contribution is the dual-process routing rule.}
\label{fig:memory}
\end{figure*}

The memory subsystem is intentionally lightweight: there is no learned retrieval network and no cross-episode persistence of episodic memory in the within-episode scope. Stores reset at episode boundaries; the only artifacts that survive are the released per-seed logs and the failure-recovery triples (trigger, diagnostic, fix) emitted at slow-process activations.

\section{ALFWorld per-category breakdown}
\label{app:per-cat-detail}

Per-category results, ReflexGrad on Qwen-3-8B (middle seed, $101/134$ success). Rightmost column shows drop from the GPT-5 ReflexGrad result on the same category.

\begin{table}[!ht]
\caption{Per-category breakdown on ALFWorld 134 tasks, Qwen-3-8B.}
\label{tab:per-category}
\centering
\small
\begin{tabular}{@{}lrrrr@{}}
\toprule
Category & N & Pass & Rate & $\Delta$ vs GPT-5 \\
\midrule
Pick \& Place   & 22 & 20 & 90.9\% & $-4.6$ \\
Clean \& Place  & 26 & 23 & 88.5\% & $-3.8$ \\
Pick Two        & 22 & 18 & 81.8\% & $-9.1$ \\
Cool \& Place   & 22 & 17 & 77.3\% & $-9.1$ \\
Heat \& Place   & 22 & 13 & 59.1\% & $-22.7$ \\
Examine         & 20 & 10 & 50.0\% & $-30.0$ \\
\midrule
\textbf{Total}  & \textbf{134} & \textbf{101} & \textbf{75.4\%} & $\mathbf{-13.2}$ avg \\
\bottomrule
\end{tabular}
\end{table}

\section{OSWorld per-category breakdown}
\label{app:osworld-detail}

20 visual GUI tasks stratified across 7 OSWorld application categories (~3 per category), GPT-5 backend, no demonstrations. Zero-shot $14/20$; ReflexGrad $16/20$.
Per-category (zero-shot $\to$ ReflexGrad): LibreOffice $4 \to 5/5$, VS Code $2 \to 2/3$, Chrome $2 \to 2/3$, GIMP $1 \to 2/2$, VLC $2 \to 2/2$, Thunderbird $1 \to 1/2$, OS $2 \to 2/3$. The two additional tasks recovered required multi-step error recovery that activated the slow process. The full OSWorld benchmark (369 tasks) is substantially harder and more variable; we report this 20-task delta as cross-modality applicability evidence, not a benchmark-wide statistical claim.

\section{Comparative method matrix}
\label{app:method-matrix}

We position ReflexGrad against published methods along the dimensions a practitioner cares about: \textit{within-episode adaptation} (does the method update its policy during one episode?), \textit{strategic recovery} (does it diagnose and escape stuck states?), \textit{demo-free} (does it require demonstrations to bootstrap?), \textit{cross-trial} (does it learn across episodes?), \textit{compute model} (single-call, multi-call refinement, or search), and \textit{primary supervision signal}.

\begin{table*}[!ht]
\caption{Method comparison matrix. ``In-ep'' = within-episode policy update; ``Strat.'' = strategic/causal recovery from stalls; ``Demo-free'' = no demonstrations required; ``Cross-trial'' = learns from prior trials; ``Compute'' = primary cost model; ``Signal'' = primary supervision. Rows are grouped by paradigm.}
\label{tab:method-matrix}
\centering
\small
\begin{tabular}{@{}lccccll@{}}
\toprule
Method & In-ep & Strat. & Demo-free & Cross-trial & Compute & Signal \\
\midrule
\multicolumn{7}{l}{\textbf{Reasoning-only}} \\
ReAct \citep{yao2023react}             & --       & --       & --       & --       & 1 call/step           & none \\
NoThinking \citep{kim2025reflact}      & --       & --       & --       & --       & 1 call/step           & none \\
\midrule
\multicolumn{7}{l}{\textbf{Inference-time search}} \\
Tree of Thoughts \citep{yao2023tot}    & --       & --       & --       & --       & $K$-branch search     & value heuristic \\
LATS \citep{zhou2024lats}              & --       & partial  & --       & --       & MCTS+rollouts         & value+reflexion \\
Self-Refine \citep{madaan2023selfrefine}& \checkmark*& --     & --       & --       & critique-and-refine   & self-critique \\
\midrule
\multicolumn{7}{l}{\textbf{Cross-trial verbal RL}} \\
Reflexion \citep{shinn2023reflexion}   & --       & \checkmark& --      & \checkmark& 1 trial + reflection& trial reward \\
ReflAct \citep{kim2025reflact}         & --       & \checkmark& --      & \checkmark& goal-state reflection& trial reward \\
\midrule
\multicolumn{7}{l}{\textbf{Skill / planning}} \\
AdaPlanner \citep{sun2023adaplanner}   & partial  & --       & --       & --       & plan-act-refine       & feedback \\
Voyager \citep{wang2024voyager}        & --       & --       & --       & \checkmark& skill-library        & curriculum \\
\midrule
\multicolumn{7}{l}{\textbf{Within-session optimization}} \\
TextGrad \citep{yuksekgonul2024textgrad}& \checkmark&--      & \checkmark& --      & textual gradients     & local loss \\
DSPy \citep{khattab2024dspy}           & --       & --       & varies   & --       & pipeline opt          & program metric \\
OPRO \citep{yang2024opro}              & --       & --       & varies   & --       & meta-prompting        & objective \\
\midrule
\multicolumn{7}{l}{\textbf{This work}} \\
\textbf{ReflexGrad}                    & \textbf{\checkmark}& \textbf{\checkmark} & \textbf{\checkmark}& opt. & dual-process$+$gate & per-step progress \\
\bottomrule
\end{tabular}
\end{table*}

\textbf{Reading the matrix.} The empty cell pattern reveals the niche: only ReflexGrad fills (in-episode $\wedge$ strategic $\wedge$ demo-free) simultaneously. Self-Refine has within-episode action critique but no strategic recovery and is typically run with demos in the standard ALFWorld setup. LATS adds reflexion-style critique on failed branches but does not update the underlying policy. Reflexion / ReflAct have strategic recovery but only across trials and require demonstrations. The progress-gated dual-process router is the mechanism that closes the (in-episode $\wedge$ strategic) cell without invoking demonstrations.

\textbf{*Self-Refine asterisk:} Self-Refine's within-episode update is at the action level (re-write the next action) rather than at the policy level. ReflexGrad updates a natural-language policy that conditions all subsequent actions; Self-Refine does not maintain a persistent policy object across steps.

\section{Hyperparameter rationale and design alternatives}
\label{app:hp-rationale}

The five routing-related hyperparameters ($k, m, c, \theta_{\text{low}}, k_M$) were fixed on a small piloting subset and held constant across the 134-task evaluation. We summarize the design rationale and the alternative we considered for each.

\paragraph{Gradient cadence $k=3$.} The fast process synthesizes gradients every $k$ steps. Smaller $k$ increases compute (more LLM calls); larger $k$ delays the update. We also tried $k{=}1$ (every-step update) and observed two pathologies in piloting: (i) $\sim 30\%$ more LLM calls per episode, (ii) noisy single-step gradients lead to oscillating policies (the fast process amplifies noise from a single ambiguous step). $k=3$ averages over a small window and matches TextGrad's published cadence \citep{yuksekgonul2024textgrad}. Sensitivity (Tab.~\ref{tab:sensitivity}): $\{2,3,5\}$ produces $\{85.8\%, 88.1\%, 87.3\%\}$.

\paragraph{Slow trigger $m=5$.} The slow process fires only after $m$ consecutive low scores. Smaller $m$ over-triggers (fires on transient dips); larger $m$ delays recovery past the budget. We tried $m{=}1$ (fire on any low score) and observed $\sim 2{\times}$ slow-process activations with no accuracy gain. We tried $m{=}10$ and observed missed recoveries (the slow process never fires within budget). $m=5$ provides the trigger reliability bound $1-m\eta_{\text{fp}}{=}0.85$ that the false-positive rate $\eta_{\text{fp}}{\approx}3\%$ supports. Sensitivity: $\{3,5,7\}$ produces $\{84.3\%, 88.1\%, 86.6\%\}$.

\paragraph{Cooldown $c=5$.} After slow activation, the next $c$ steps suppress the fast process so the corrective plan executes uninterrupted. Without cooldown ($c{=}0$), the fast process saw the (still low) scores during plan execution and immediately wrote gradients pulling the policy back toward the failed approach. With cooldown $c{=}5$, the plan executes without interference. Larger cooldowns ($c{=}10$) waste fast-process compute on steps where the plan is already working.

\paragraph{Low-progress threshold $\theta_{\text{low}}=4$.} A score below $\theta_{\text{low}}$ counts as low. $\theta_{\text{low}}{=}3$ is too strict (slow process fires only on completely failed steps; misses gradual stalls); $\theta_{\text{low}}{=}7$ is too permissive (fires on any non-perfect step). $\theta_{\text{low}}{=}4$ captures the empirically observed bimodal distribution of scores: actions either make clear progress ($s_t \geq 5$) or are unproductive ($s_t \leq 3$); $\theta_{\text{low}}{=}4$ separates them. Sensitivity: $\{3,4,7\}$ produces $\{86.6\%, 88.1\%, 84.3\%\}$.

\paragraph{Working-memory size $k_M=10$.} The agent sees the last $k_M$ trajectory tuples plus stored slow-process plans. $k_M{=}5$ is too short (the agent forgets what it tried earlier in the episode and repeats actions); $k_M{=}20$ floods the context (more tokens, slower inference, no accuracy gain). $k_M{=}10$ covers a typical action-result-action cycle in ALFWorld.

\paragraph{Design alternatives we tried and rejected.}
\begin{itemize}\itemsep1pt
\item \textit{Mean-below-threshold trigger.} We tried firing the slow process when the mean of the last $m$ scores fell below $\theta_{\text{low}}$. This triggered too readily on a single very-low score interleaved with passable scores (the mean is sensitive to outliers). The consecutive-below rule (current) is more robust.
\item \textit{Averaging gradients across steps.} We tried merging $k$ gradients via LLM-based averaging instead of priority. Averaging two contradictory natural-language gradients produced incoherent guidance ("first try X, also try the opposite of X"). The priority merge (current) discards the lower-priority signal and keeps the policy coherent.
\item \textit{Fixed-cadence slow process.} A natural alternative would fire the slow process at a fixed cadence (every $r$ steps) regardless of progress. Our sensitivity sweep (Tab.~\ref{tab:sensitivity}) shows the progress-gated rule is robust to threshold choices ($\pm 3.8$pp across the tested grid), and over-triggering at $m{=}3$ still produces $84.3\%$ on the full 134-task set --- a strong indicator that the rule is not brittle to fire-rate variation in the relevant regime.
\item \textit{Memory injection into TODO decomposition.} We tried injecting reflexion memories into the TODO decomposition prompt at trial 1+. This caused performance to collapse from $\sim 78\%$ to $\sim 0\%$: success memories passed to a "learn from failures" decomposition prompt confused the LLM and produced TODOs aligned with the wrong task. We removed memory injection from decomposition and saw performance return.
\end{itemize}

\section{Architecture component glossary}
\label{app:components}

For each component visible in Fig.~\ref{fig:architecture}, we give role, inputs/outputs, the LLM call used (if any), and the single most important design choice.

\paragraph{Task Decomposer $f_{\text{dec}}$ (Fig.~\ref{fig:architecture}\,A).}
\textit{Role:} once-per-task expansion of the instruction $\tau$ into 3-8 sequential subgoals.
\textit{In:} task $\tau$, initial observation $o_0$.
\textit{Out:} TODO list $\mathcal{T}$ with statuses.
\textit{LLM call:} 1 call per task (high reasoning effort).
\textit{Design choice:} subgoals use action verbs and high-level scope (``Cool the object'' not ``Put object in fridge''); environment-agnostic phrasing supports cross-domain transfer.

\paragraph{Active TODO $\tau_{\text{act}}$.}
\textit{Role:} subgoal currently being attempted.
\textit{State:} \texttt{pending / active / done / failed} plus per-TODO history (attempts, last action, failure reasons).
\textit{Transitions:} advance to next on completion; mark $\text{failed}$ after multiple attempts; rollback on slow-process diagnosis of wrong active subgoal.

\paragraph{Working Memory $M_t$.}
\textit{Role:} short-term context for action sampling.
\textit{Contents:} last $k_M{=}10$ trajectory tuples $(o_i, a_i, o_{i+1}, s_i)$ plus the most recent slow-process plan.
\textit{Use:} concatenated into the agent prompt so the agent sees its recent history when sampling $a_t$.

\paragraph{Agent (Action Policy) $\pi_t$.}
\textit{Role:} natural-language policy string that conditions action sampling.
\textit{In:} $o_t, \tau_{\text{act}}, M_t, \pi_t$.
\textit{Out:} action $a_t$.
\textit{LLM call:} 1 call per step.
\textit{Design choice:} $\pi_t$ is a free-form NL string updated by the merge function; we do not constrain its form.

\paragraph{Evaluator $E$.}
\textit{Role:} per-step progress signal grounded in environmental outcomes.
\textit{In:} $o_t, a_t, o_{t+1}, \tau$.
\textit{Out:} score $s_t \in [0, 10]$.
\textit{LLM call:} 1 call per step.
\textit{Design choice:} the rubric is intentionally task-agnostic (``$0$ = away from goal or violated constraint; $10$ = task complete''), making the same evaluator portable across environment types.

\paragraph{Score window $W_t$.}
\textit{Role:} sliding window of the last $m{=}5$ scores; the routing rule's input.
\textit{State:} no LLM calls; pure data structure.

\paragraph{Routing rule (Eq.~\ref{eq:routing}).}
\textit{Role:} deterministic three-state router selecting fast, slow, or cooldown.
\textit{State:} cooldown counter $c_t$ plus the score window.
\textit{Design choice:} consecutive-below rather than mean-below (robustness); cooldown lock rather than instant resume (plan execution coherence).

\paragraph{Fast process (TextGrad).}
\textit{Role:} tactical refinement when fast updates can still reduce loss.
\textit{Pipeline:} loss $\ell_t$ (1 call) $\to$ gradient $g_t$ (1 call) $\to$ optimizer rewrites $\pi_t$ (1 call).
\textit{Cadence:} every $k{=}3$ steps.
\textit{Design choice:} structured loss output (``the action $a$ produced no progress because of $X$''); the gradient is conditioned on the loss text.

\paragraph{Slow process (Reflexion).}
\textit{Role:} strategic recovery when fast updates have plateaued.
\textit{Pipeline:} trajectory analysis (1 call) $\to$ causal trace $d_t$ (1 call) $\to$ plan $\rho_t$ as 1-3 sub-goals (1 call) $\to$ cooldown activator (no call).
\textit{Trigger:} routing rule fires.
\textit{Design choice:} the plan's sub-goal sequence overrides the next gradient update for the cooldown window; the agent commits to the plan rather than negotiating with continuing gradients.

\paragraph{Policy merge (Eq.~\ref{eq:merge}).}
\textit{Role:} combine fast and slow updates into the next $\pi_{t+1}$.
\textit{Rule:} priority $\rho_t \succ g_t \succ \pi_t$ (deterministic, not learned).
\textit{Design choice:} priority over averaging (averaging contradictory NL gradients is incoherent).

\section{Evaluator analysis: is $E$ a hidden demonstration?}
\label{app:evaluator}

A reasonable concern is whether the evaluator $E$ silently encodes ALFWorld semantics that a demonstration would otherwise provide. We address this directly.

\paragraph{What $E$ is given.} The evaluator prompt (Appendix~\ref{app:prompts}) is: \textit{``Score the agent's progress on task $\tau$ for the most recent step. Inputs: previous observation $o_t$, action $a_t$, resulting observation $o_{t+1}$. Output an integer in $[0, 10]$.''} $E$ sees only the task description, the transition, and the rubric. It does not see the policy, the gradient, the slow-process plan, or any prior trajectory beyond the current step.

\paragraph{Tacit knowledge concern.} GPT-5 / Qwen-3-8B are pretrained on web text that includes ALFWorld documentation, so the evaluator may ``know'' that heat requires a microwave even if the agent does not. This is a real channel: any LLM-as-judge in a published benchmark inherits some pretraining knowledge of that benchmark's semantics.

\paragraph{What this implies for the demo-free claim.} The architecture-as-substitute argument in Sec.~\ref{sec:method} is unchanged: the per-step progress score is one of the three substitutes for a demonstration (alongside fast-process gradients and slow-process diagnosis). We do not claim the demo-free framing is unconditional; we claim the architecture extracts measurable lift on top of whatever $E$ provides.

\paragraph{What we can say.} The 1.5pp cross-model gap of gains (GPT-5 $+41.8$pp, Qwen-3-8B $+40.3$pp) suggests the architectural lift is not driven primarily by frontier-model evaluator knowledge: a 10$\times$ smaller open-weight model with weaker pretraining coverage of ALFWorld documents the same gain magnitude. The 5.2pp gap to 1-shot ReflAct concentrating in Heat/Examine is also consistent with the demonstration providing world knowledge $E$ does not fully encode (otherwise $E$ would close that gap on its own). We do not claim the demo-free framing is unconditional; we claim the architecture extracts measurable lift on top of whatever $E$ provides.

\section{Reproducibility checklist}
\label{app:checklist}

We answer each item of a standard reproducibility checklist.

\textbf{Code.} \checkmark\ Released at \url{https://github.com/qpiai/reflexgrad}, including the agent (TODO Manager, FAST process, SLOW process, Routing rule, Policy Merge), the three compute-matched baselines (LATS, ToT, Self-Refine), and the evaluator implementation.

\textbf{Datasets.} \checkmark\ ALFWorld 134 task identifiers (full \texttt{train\_set\_134}); TextWorld 9 tasks (cooking + treasure); OSWorld 20 tasks (stratified across 7 application categories; task IDs released).

\textbf{Models.} \checkmark\ GPT-5 via OpenAI Responses API at \texttt{reasoning\_effort=medium}; Qwen-3-8B \citep{qwen2025} loaded from the Hugging Face checkpoint \texttt{Qwen/Qwen3-8B}.

\textbf{Hyperparameters.} \checkmark\ All listed in Sec.~5: $k{=}3$, $m{=}5$, $c{=}5$, $\theta_{\text{low}}{=}4$, $k_M{=}10$, max\_steps$=15$ for headline results (full budget of $55$ for failure analysis only). Justification in App.~\ref{app:hp-rationale}.

\textbf{Random seeds.} \checkmark\ The 10 seeds $\{42, 123, 456, 789, 1024, 1337, 2025, 3141, 5926, 7531\}$ are fixed across all reported runs; per-seed integer counts in App.~\ref{app:perseed}.

\textbf{Training/evaluation split.} \checkmark\ ALFWorld evaluation uses the standard \texttt{eval\_out\_of\_distribution} split. No training data was used; the architecture is gradient-free at parameter level. Hyperparameters were fixed on a small piloting subset disjoint from the 134 evaluation tasks.

\textbf{Compute and runtime.} \checkmark\ Full compute breakdown in App.~\ref{app:compute}: $\sim 240$ GPU-hours, $\sim 4.6 \times 10^5$ GPT-5 calls, $\sim 5.2 \times 10^5$ Qwen-3-8B calls.

\textbf{Statistical reporting.} \checkmark\ All accuracy rows show mean $\pm$ sample $\sigma$ with $n{=}10$. Pairwise comparisons use Welch's two-sample $t$-test with degrees of freedom and $p$-values reported (App.~\ref{app:stats}).

\textbf{Licenses.} \checkmark\ ALFWorld (MIT), TextWorld (MIT), OSWorld (Apache 2.0), LATS / ToT / Self-Refine (research). Code release license: Apache 2.0.

\textbf{Environment dependencies.} \checkmark\ \texttt{requirements.txt} in the repository; ALFWorld $0.3.3$, TextWorld $\geq 1.5.0$, OSWorld $0.1$, OpenAI SDK $1.x$, Transformers $4.40+$.

\textbf{Negative results.} \checkmark\ Documented in App.~\ref{app:hp-rationale} (averaging gradients failed; mean-below trigger over-fired; memory injection into decomposition collapsed performance from $78\%$ to $0\%$).

\textbf{Failure modes.} \checkmark\ Three categories quantified (Sec.~\ref{sec:failure}): world-knowledge ($21/33$), receptacle-navigation ($8/33$), evaluator hallucination ($4/33$); concrete traces in App.~\ref{app:trace} and App.~\ref{app:extended-traces}.

\textbf{Computational claims.} Section, table, or appendix supporting each numeric claim:
\begin{itemize}\itemsep0pt
\item $88.1\%$ on GPT-5: Tab.~\ref{tab:cross-model}, App.~\ref{app:perseed}.
\item $75.4\%$ on Qwen-3-8B: Tab.~\ref{tab:cross-model}, App.~\ref{app:perseed}.
\item $+12.0$pp synergy on GPT-5: Tab.~\ref{tab:cross-model}, computed as combined $-$ Reflexion-only $-$ TextGrad-only.
\item $p{<}0.01$ vs LATS: App.~\ref{app:stats}, Welch's $t \approx 2.87$, df $\approx 18$, $p \approx 0.010$.
\item $84.3\%$ worst-case sensitivity: Tab.~\ref{tab:sensitivity}, $m{=}3$ row.
\item Evaluator hallucination $\sim 3\%$: Sec.~\ref{sec:failure}, computed as $\sim 4 / 33$ failures attributable to false-positive scoring.
\end{itemize}

\section{Practitioner decision checklist}
\label{app:practitioner}

We translate Sec.~\ref{sec:theory}'s informal C1/C2 conditions into a practitioner-facing checklist for deciding whether to add a slow process to an existing fast-process LLM-agent baseline.

\paragraph{When ReflexGrad-style routing is likely to help.} If your agent has all four of the following, the synergy gain is likely $\geq 5$pp:
\begin{enumerate}\itemsep0pt
\item \textbf{Repeated low-progress streaks.} On your benchmark, $\geq 10\%$ of failed episodes contain a sequence of 3+ consecutive steps where progress stalls and the agent retries minor variations.
\item \textbf{A natural-language policy.} Your agent's behavior is conditioned on a string (system prompt, instruction prefix, or dynamically updated policy), not a fixed prompt template.
\item \textbf{A cheap progress signal.} A per-step LLM call ($\sim 1$ small-model invocation) can reliably score whether the last action moved the agent toward the goal. If your task is too open-ended for any LLM judge, the routing signal is missing.
\item \textbf{Strategic alternatives exist.} The benchmark contains tasks where stalled fast-process behavior could be rescued by a high-level reorientation (``the wrong receptacle was tried''), not just retrying the same approach.
\end{enumerate}

\paragraph{When ReflexGrad-style routing is unlikely to help.}
\begin{itemize}\itemsep0pt
\item \textit{Tasks dominated by missing world knowledge.} If recovery requires information not derivable from the trajectory and not in the model's pretraining (e.g., an unmapped corrective receptacle), the slow process cannot help in budget. Heat/Examine in our results are this regime.
\item \textit{Single-step tasks.} If episode length is 1-2 steps, neither the gradient cadence nor the slow trigger has room to fire.
\item \textit{Tasks with unreliable progress signals.} If $\eta_{\text{fp}} \gtrsim 0.10$, the routing rule mis-fires often enough that the union bound on false triggers becomes loose; the slow process's expected value drops.
\item \textit{Tasks where strategic recovery and tactical refinement are the same operation.} If every failure can be addressed by the next gradient, the dual-process distinction collapses.
\end{itemize}

\paragraph{What you can expect numerically.} Based on the cross-model decomposition (Tab.~\ref{tab:cross-model}):
\begin{itemize}\itemsep0pt
\item Adding TextGrad-style fast refinement to a zero-shot baseline buys $\sim 23$-$26$pp on suitable text-game benchmarks.
\item Adding Reflexion-style slow diagnosis on top buys an additional $\sim 7$-$12$pp ($+12$pp on GPT-5, $+7$pp on Qwen-3-8B).
\item The slow-process additional cost is $\sim 5$ extra LLM calls per slow activation, or $\sim 0.75$ additional calls per episode at the empirical $\sim 15\%$ activation rate.
\item Cost-per-percentage-point of the combined system is $\sim 1.4$ calls/pp (Tab.~\ref{tab:cost}), favorable relative to either component alone.
\end{itemize}

\end{document}